\documentclass[lettersize,journal,twocolumn]{IEEEtran}

\usepackage{mathptmx} 
\usepackage{times} 
\usepackage[dvipdfmx]{graphicx}
\usepackage{url}
\usepackage{graphicx}
\usepackage[figurename=Fig.\ ]{caption}
\usepackage{epsfig} 
\usepackage{cite}
\usepackage{color}
\usepackage{booktabs}  %
\usepackage{amsmath}
\usepackage{amssymb}
\usepackage{indentfirst}
\usepackage{algorithm}
\usepackage{algorithmicx}
\usepackage{algpseudocode}
\usepackage{caption}
\usepackage{subcaption}
\usepackage{comment}
\usepackage{color,soul}
\usepackage{ifthen}
\usepackage[table,xcdraw]{xcolor}
\usepackage{listings}
\begin{document}

\title{Applying Learning-from-observation to household service robots: three common-sense formulations}

\author{
Katsushi Ikeuchi$^*$, 
Jun Takamatsu,
Kazuhiro Sasabuchi \\
Naoki Wake,
Atsushi Kanehira
\thanks{
$^*$katsuike@microsoft.com }

\thanks{Microsoft Applied Robotics Research, Redmond WA 98052, USA}

}


\maketitle
\begin{abstract}
Utilizing a robot in a new application requires the robot to be programmed at each time. To reduce such programmings efforts, we have been developing ``Learning-from-observation (LfO)'' that automatically generates robot programs by observing human demonstrations. One of the main issues with introducing this LfO system into the domain of household tasks is the cluttered environments, which cause difficulty in determining which elements are important for task execution when observing demonstrations. To overcome this issue, it is necessary for the system to have common sense shared with the human demonstrator. 
This paper addresses three relationships that LfO in the household domain should focus on when observing demonstrations and proposes representations to describe the common sense used by the demonstrator for optimal execution of task sequences. Specifically, the paper proposes to use labanotation to describe the postures between the environment and the robot, contact-webs to describe the grasping methods between the robot and the tool, and physical and semantic constraints to describe the motions between the tool and the environment.
Then, based on these representations, the paper formulates task models, machine-independent robot programs, that indicate what to do and how to do. Third, the paper explains the task encoder to obtain task models and task decoder to execute the task models on the robot hardware. Finally, this paper presents how the system actually works through several example scenes.

\end{abstract}

\section{Introduction}
One powerful means of acquiring human behavior is to observe and imitate the behavior of others. Humans go through a period of imitating their mother's behavior at one stage of development~\cite{piaget2020psychologie}. Even in adulthood, imitation from practice videos is often used in various sport practice such as golf practice and judo practice. Learning-from-observation, programming-by-demonstration and learning-by-watching aim to apply this behavior observation and learning paradigm to robots and to automatically generate robot programs from observation~\cite{schaal1999imitation,schaal2003imitation,billard2008robot,asfour2008imitation,dillmann2010advances,akgun2012keylfd}. The origin of this field lies in our research~\cite{ikeuchireddy1991,ikeuchi1994toward} as well as Kuniyoshi~\cite{kuniyoshi1994learning}. These two studies shared the common goal that they attempted to understand human behavior by viewing it under some framework and to make robots to perform the same behavior following the framework.

Later, in terms of the approach to obtain this framework for observation, the research field was split into two schools: the bottom-up and the top-down. The ``bottom-up'' school~\cite{samejima2006multiple,schaal1996learning,billard2008robot} has been attempting to acquire this framework from scratch through learning. The ``top-down" school~\cite{kang1997toward,tsuda2000generation,takamatsu2006representation} has been attempting to mathematically design a framework by utilizing the accumulated knowledge of robotics field to date. The ``bottom-up" approach has been the mainstream due to the population of researchers and the rise of machine learning.

The authors, on the other hand, take the ``top-down" position. In both human and robot learning, the physical structure, height, and weight of the student often differ from those of the teacher. Furthermore, environmental changes are also present.
Therefore, trying to mimic the teacher's trajectory itself, as in the bottom-up approach, is difficult due to these differences in kinematics, dynamics, and environment. This difficulty is avoided by first extracting the essence of the behavior from the observation and then mapping this essence in the way adapted to the individual hardware. For this purpose, we aim to design mathematically consistent abstracted behavioral models based on the knowledge accumulated in the robotics field, and to utilize these models to represent the essence of the demonstrations.

Going back to the history of robot programming, since its early days, some researchers have worked on automatic programming, to generate robot programs from assembly drawings or abstract concepts, referred to as task descriptions~\cite{lozano1983robot,lozano1984automatic,de1990and}. This trend, after about 30 years of research, encountered more difficult obstacles than imagined, as summarized in Raj Reddy's Turing Award talk ``AI: To dream possible dreams" in 2000~\cite{reddy2007dream}. In this talk, Reddy advocates 80\% AI to overcome this barrier. In other words, for the relatively easy part of 80\% or so, the solution is automatically obtained using the accumulated methods of automated programming to date. The remaining 20\% or so of the problem, which cannot be solved by any means, should be solved with hints from humans.

Our ``top-down'' approach is based on this 80\% AI approach. It attempts to overcome the difficulties of automated programming by designing a framework of understanding based on the theory accumulated in previous automated programming efforts in the field of robotics, and by gleaning hints from human demonstrations to actually make the system work.


This paper focuses on the ``top-down" approach, provides some findings on this approach, and describes some ongoing projects to apply these findings to the development of household robots. In the next section, we review previous research along with the LfO paradigm. Section 3 describes the common sense that must be shared between human demonstrators and the observation systems in order to apply this paradigm to the household domain and discusses the formulation of this knowledge. Section 4 describes the grasping and the manipulation skill libraries necessary to implement this formulated knowledge in robots. Sections 5 and 6 describe the system implementation and the system in action. Section 7 summarizes this paper.

The contribution of this paper are:
\begin{itemize}
    \setlength{\parskip}{0pt}
    \item to give an overview of the past LfO efforts,
    \item to enumerate the three common senses required to apply LfO to the household domains,
    \item to propose representations of the three common senses and to show how these representations are used in LfO task-encoder and task-decoder.
\end{itemize}

\section{LfO paradigm}

\begin{figure}[h]

\begin{tabular}{cc}

\begin{minipage}{0.5\hsize}
\centering
\includegraphics[height=40mm]{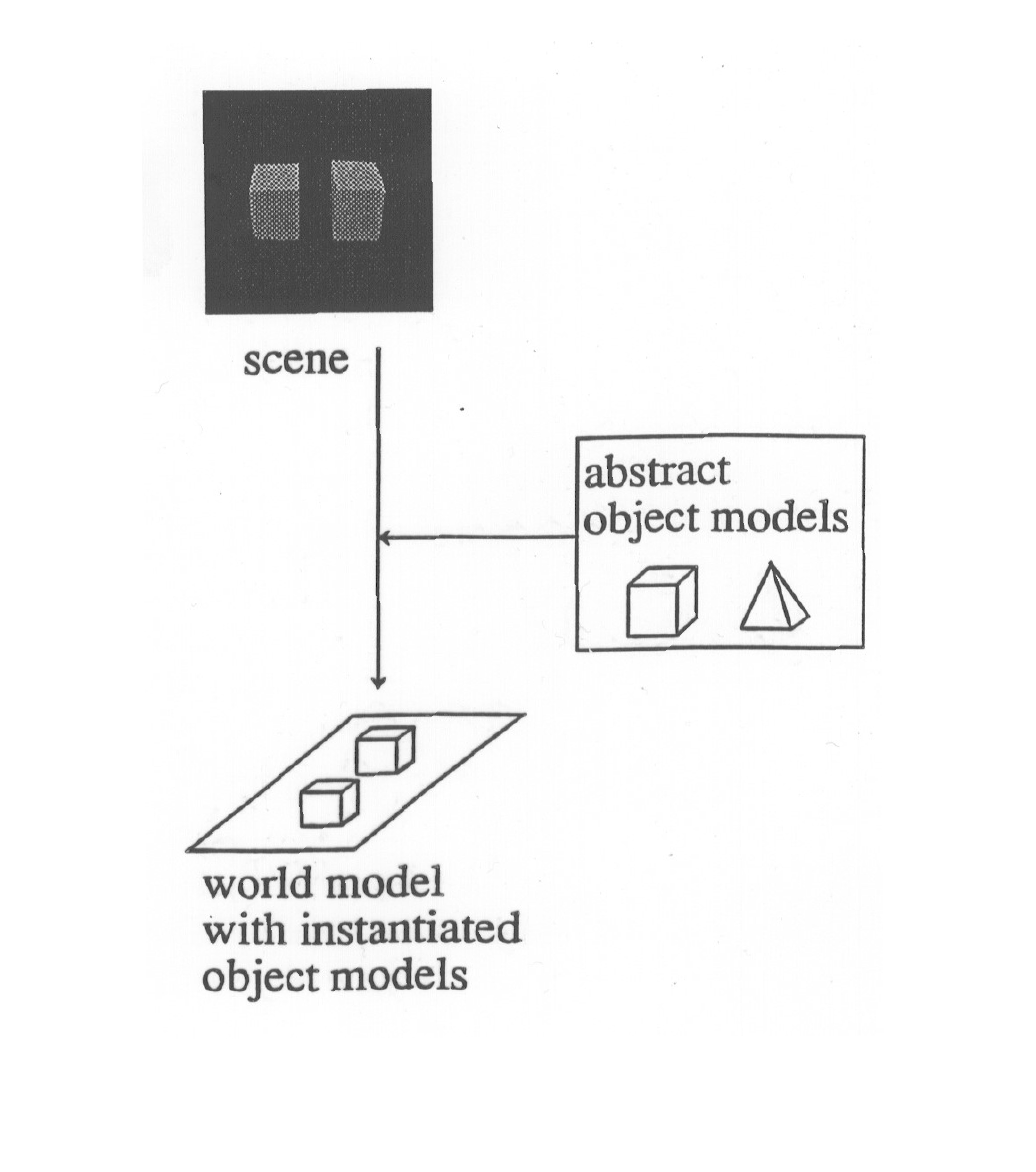} \\
(a) Object recognition
\end{minipage}

&

\begin{minipage}{0.5\hsize}
\centering
\includegraphics[height=40mm]{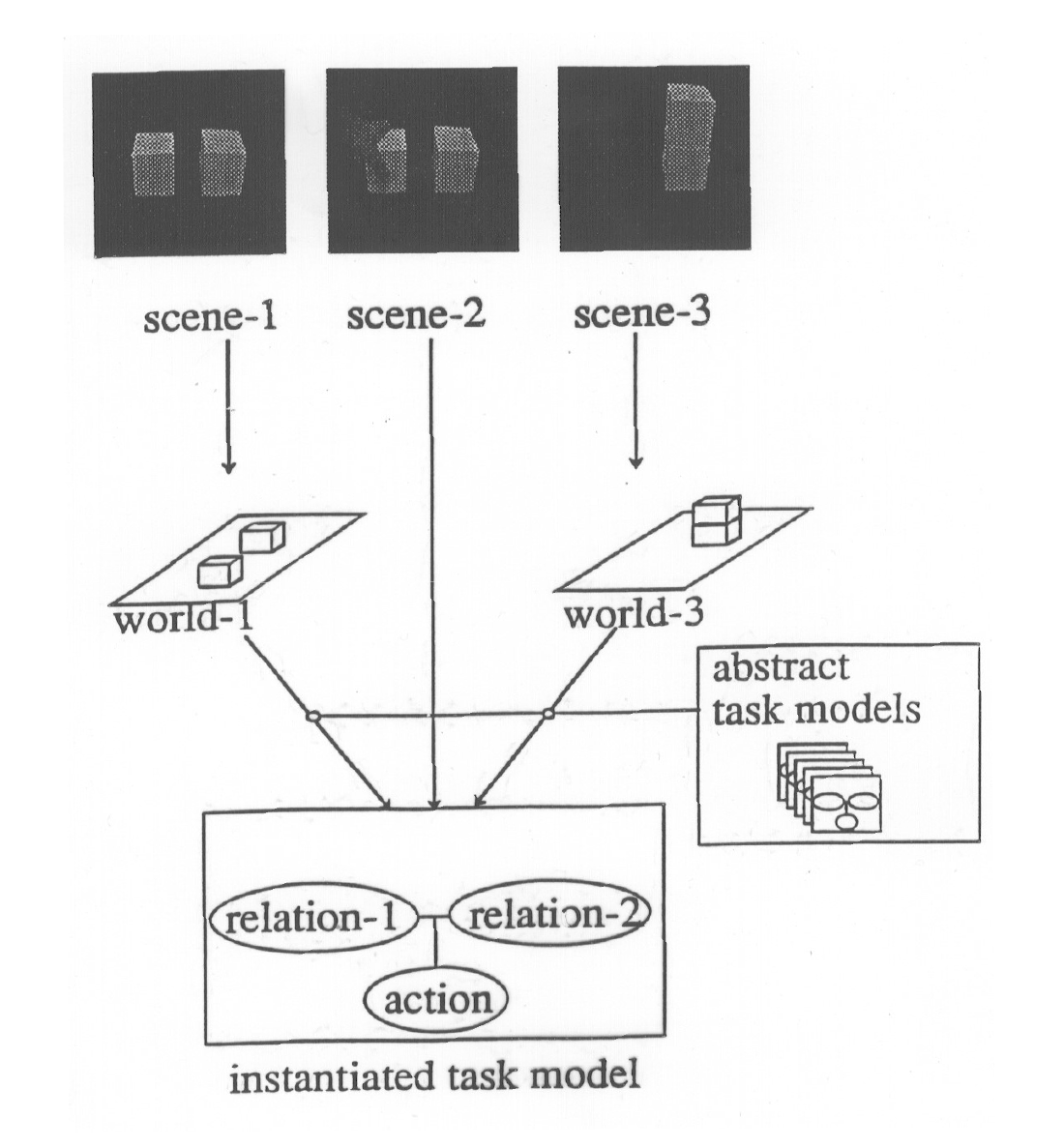} \\ 
(b) Task recognition
\end{minipage}

\end{tabular}

\caption{Object recognition and Task recognition}
\label{fig:object-recognition}
\end{figure}

The LfO paradigm can be considered as an extension of the Marr's paradigm of object recognition~\cite{marr2010vision}. First, let us review the Marr's paradigm, shown in Fig.~\ref{fig:object-recognition}(a). In this paradigm, an abstract object model is created inside the computer and then matching is performed between the internal model and the external image. The matching process involves two distinct sub-processes: indexing and localization.
Indexing is the process of identifying which of the abstract object models corresponds to the observed one. Namely, indexing obtains the solution corresponding to what this is. Localization is the process of transforming the abstract object model into an instantiated object model with concrete dimensions and placing it in virtual space. In other words, localization obtains a solution corresponding to where this is. In computer vision, these two operations are often performed sequentially. In the brain, these two sub-processes are carried out in separate and distinct circuits, with indexing proceeding in a new pathway through the visual cortex, and localization proceeding in an older pathway through the superior colliculus~\cite{ramachandran2003emerging}. In either case, a world model with instantiated object models, a copied world named by Marr, is eventually created in the computer or the brain~\cite{marr2010vision}.

Task recognition can be viewed as an extension of the Marr's object-recognition paradigm, as shown in Fig.~\ref{fig:object-recognition}(b)~\cite{ikeuchi1994toward}. An abstract task model is created that associates a state transition with an action required to cause such a transition. An abstract task model takes the form of Minsky's frames~\cite{minsky1988society}, in which the slots for the skill parameters required to perform the action are also prepared. Object recognition is then performed on the input images to create two world models before and after an action. Task recognition instantiates a particular task model corresponding to the action based on the state transition extracted by comparing the pre- and post-action worlds. Namely, the what to do is obtained. Third, the skill parameters for the action, such as the starting point, the end point, and the grasping point, are extracted from the images, and the instantiated task model is completed. Namely, the how to do is obtained. The instantiated task model guides the robot to imitate human actions.

When designing task models, a divide-and-conquer strategy is used to address the different domains of human activity. It would be difficult, if not impossible, to design a set of abstract task models that completely covers all human activity domains. It would be inefficient, too. Therefore, we divide various human activities into specific domains and design a set of task models that satisfies the necessary and sufficient conditions for complete coverage of those domains. The domains we have dealt with so far include:

\begin{itemize}
\item Two-block domain ~\cite{ikeuchireddy1991}
\item Polyhedral object domain (translation only)~\cite{ikeuchi1994toward}
\item Polyhedral object domain (Translation and rotation)~\cite{takamatsu2007recognizing}
\item Machine-parts domain including screw tightening, snap pushing, and rubber-ring hanging~\cite{sato2002task}
\item Knot-tying domain including eight-knot and bowline-knot~\cite{takamatsu2006representation} 
\item Folk dance domain~\cite{nakaoka2007learning,okamoto2014toward}
\end{itemize}

\begin{figure}[h]
    \centering
    \includegraphics[height=65mm]{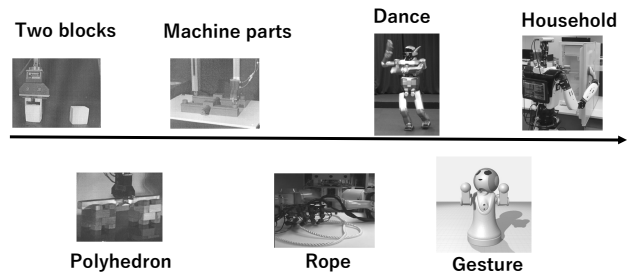}
    \caption{Domains explored}
    \label{fig:domain_explored}
\end{figure}

\section{Common-sense}
\label{section:common}

\begin{figure}[h]

\begin{tabular}{cc}

\begin{minipage}{0.5\hsize}
\centering
\includegraphics[height=25mm]{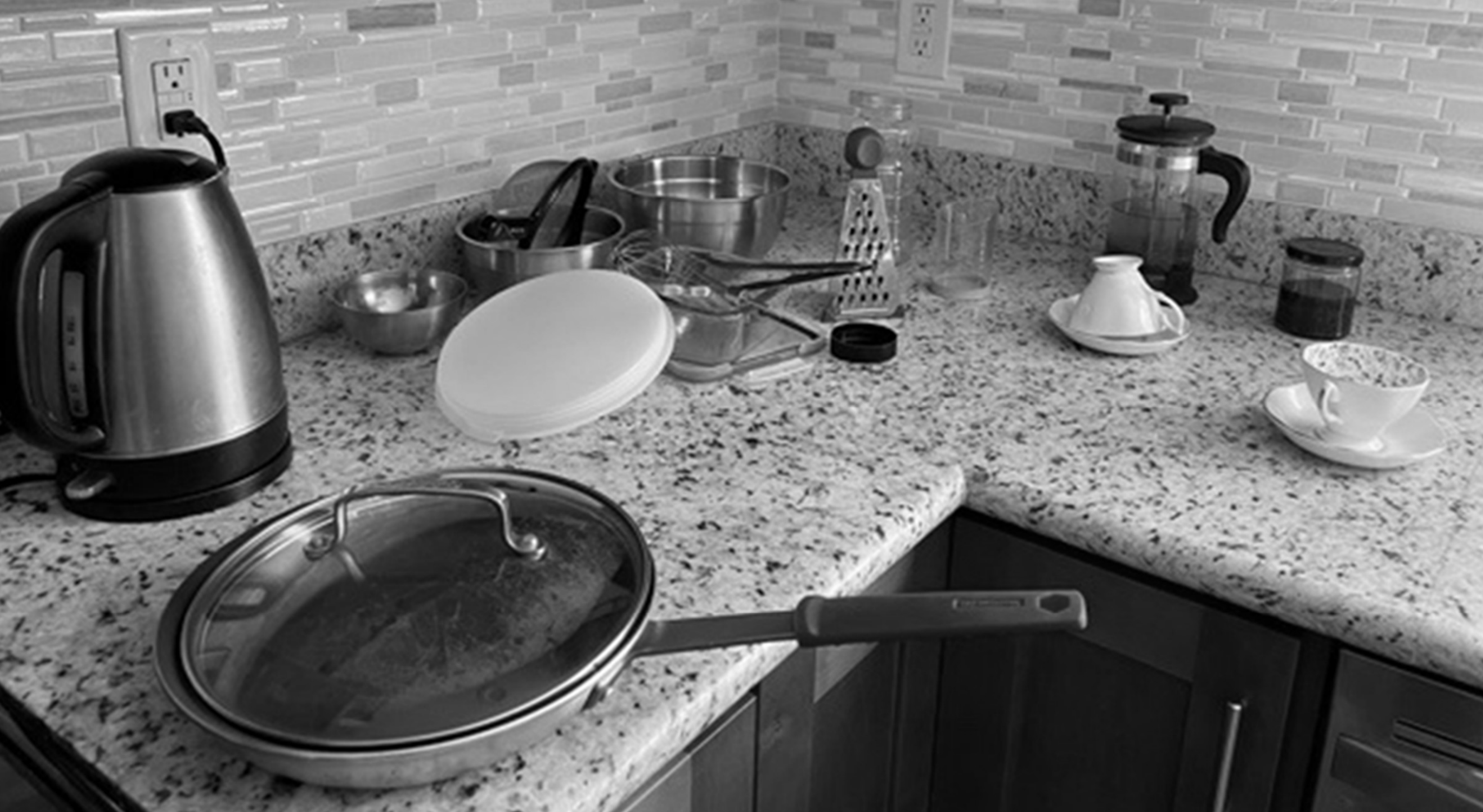} \\
(a) Clusstered environment
\end{minipage}

&

\begin{minipage}{0.5\hsize}
\centering
\includegraphics[height=25mm]{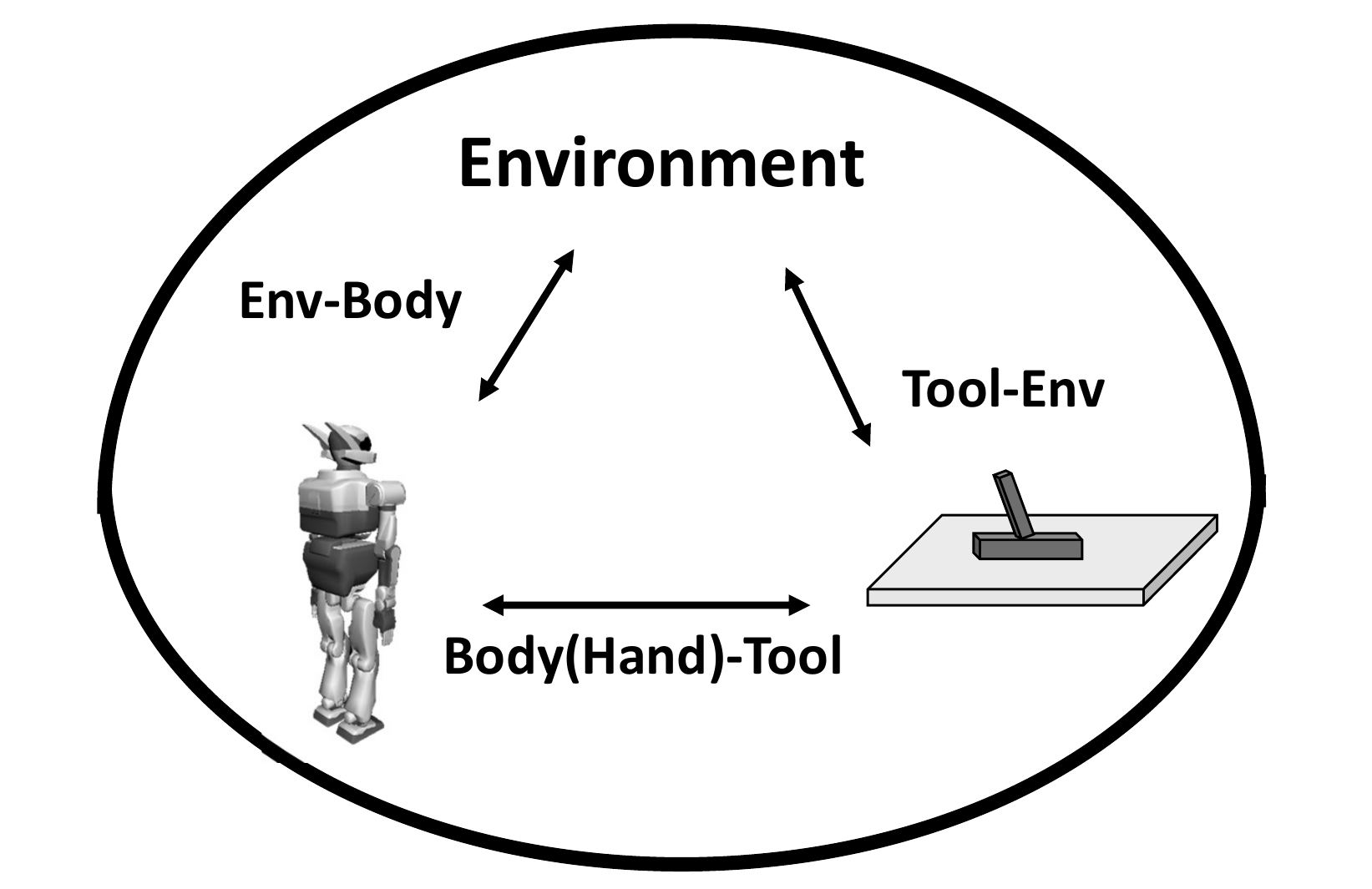} \\ 
(b) Three relations
\end{minipage}

\end{tabular}

\caption{Service-robot domain.}
\label{fig:homeEnv}
\end{figure}

When applying the LfO paradigm to the household domain in the human home environment, the main obstacle is the overwhelming crowding in this domain (See Fig.~\ref{fig:homeEnv}(a)). In the traditional industrial domain, the only objects present in the surroundings are those related to the task, to the exclusion of other miscellaneous objects. There are also not so many unrelated people in the surroundings. However, as seen in the Fig.~\ref{fig:homeEnv}(a), there are unrelated objects on the table as well as various people moving around. Therefore, it is important to share the human common sense of what is the important for the demonstration between the LfO system and the demonstrator, and let the LfO system decide where to direct its attention to avoid the crowding. We propose to focus on the following three relations as shown in Fig.~\ref{fig:homeEnv}(b) to explicitly represent important expressions derived from human common sense:
\begin{itemize}
    \item {\bf environment and robot:} to represent human common sense regarding the posture a person should take with respect to the environment for smooth execution of a task sequence
    \item {\bf robot and tool:} to represent human common sense regarding grasping strategies of objects (mainly tools) for robust execution of a task sequence
    \item {\bf tool and environment:} to represent human common sense regarding the movement of the tools in relation to the environment for successful execution of a task sequence
\end{itemize}
The following sections will discuss these relationships.

\subsection{Environment-body relation}

\begin{figure}[h]

\begin{tabular}{cc}

\begin{minipage}{0.5\hsize}
\centering
\includegraphics[height=30mm]{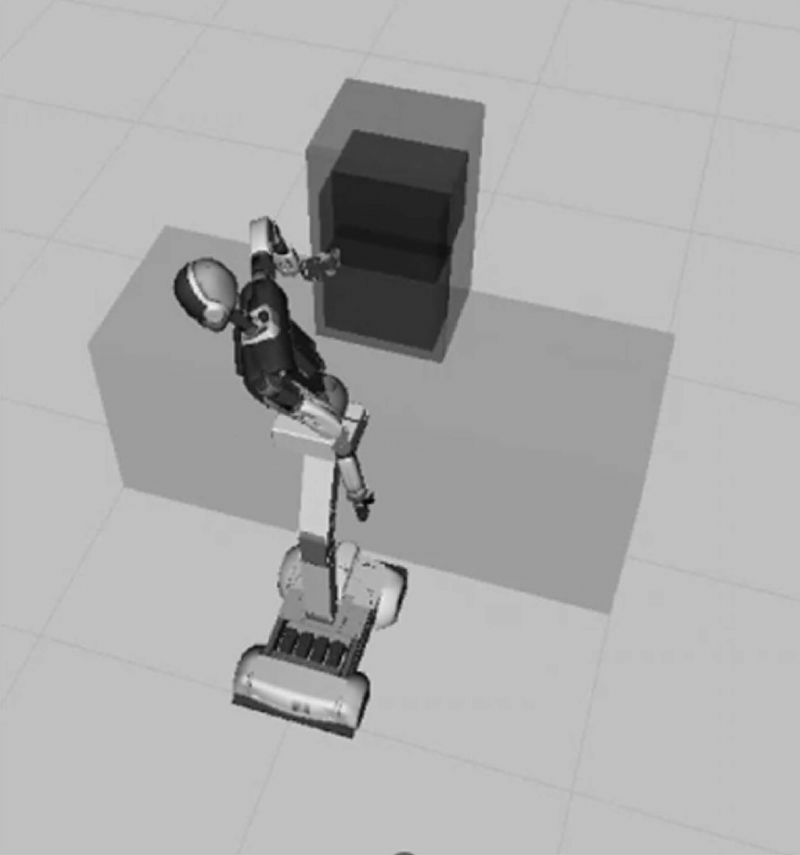} \\
(a) Inhuman-like posture
\end{minipage}

&

\begin{minipage}{0.5\hsize}
\centering
\includegraphics[height=30mm]{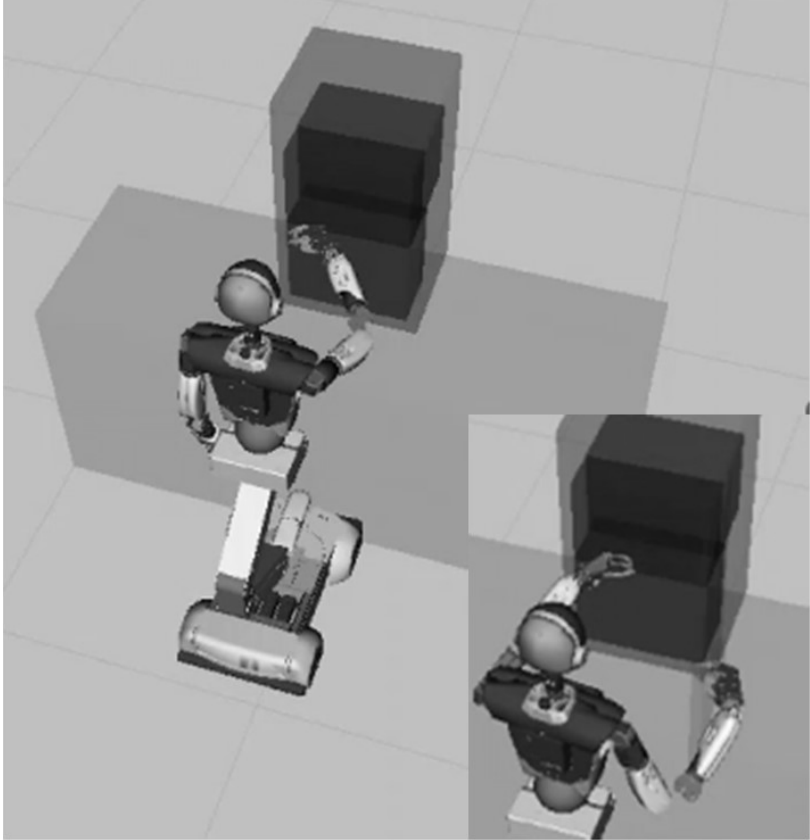} \\ 
(b) Human-like posture
\end{minipage}

\end{tabular}

\caption{Two possible postures for opening the refrigerator door.}
\label{fig:fridge}
\end{figure}

Our system is designed to mimic the approximated postures of the demonstrator. For example, when opening a refrigerator door, a robot with redundant degrees of freedom, as is common in recent humanoid robots, 
can perform this task in several different postures as shown in Fig.~\ref{fig:fridge}. However, we prefer the robot to work in a human-like posture, as shown in Fig.~\ref{fig:fridge}(b), for the following reasons:
\begin{itemize}
    \item When performing tasks, human-like postures are more predictable to bystanders and less likely to cause interpersonal accidents
    \item Humans unconsciously adopt the optimal postures for the environment in order to perform a task sequence. In the example above, the purpose of opening the door of a refrigerator is often to take out the items inside in the task. The human-like posture shown in Fig.~\ref{fig:fridge}(b) is considered more efficient.
\end{itemize}

We need a representation method to describe approximate postures.
The difference between human and robot mechanisms makes it difficult to achieve exactly the same postures by taking exactly the same joint angles and joint positions at each sampling time. Our approximate imitation does not require such precise representations. Rather, it is necessary to capture the essence of those postures that the bystanders perceive as nearly identical.

For this approximation, we will use Labanotation~\cite{GuestBOOK1970}, which is used by the dance community to record dance performances. The relationship between dance performances and Labanotation scores is similar to the relationship between music performances and music scores. Just as a piece of music can be performed from a music score, a piece of dance can be performed from a labanotation score; just as a music score can be obtained from listening to a peice of music, a Labanotation score can be obtained from watching a piece of dance. More importantly, from the same Labanotation score, each dancer, with different heights and hand lengths, performs a piece of dance that appears the same to observers. In other words, the Labanotation score is considered to capture the essence of the dance for the observers. 

\begin{figure}[h]
    \centering
    \includegraphics[height=35mm]{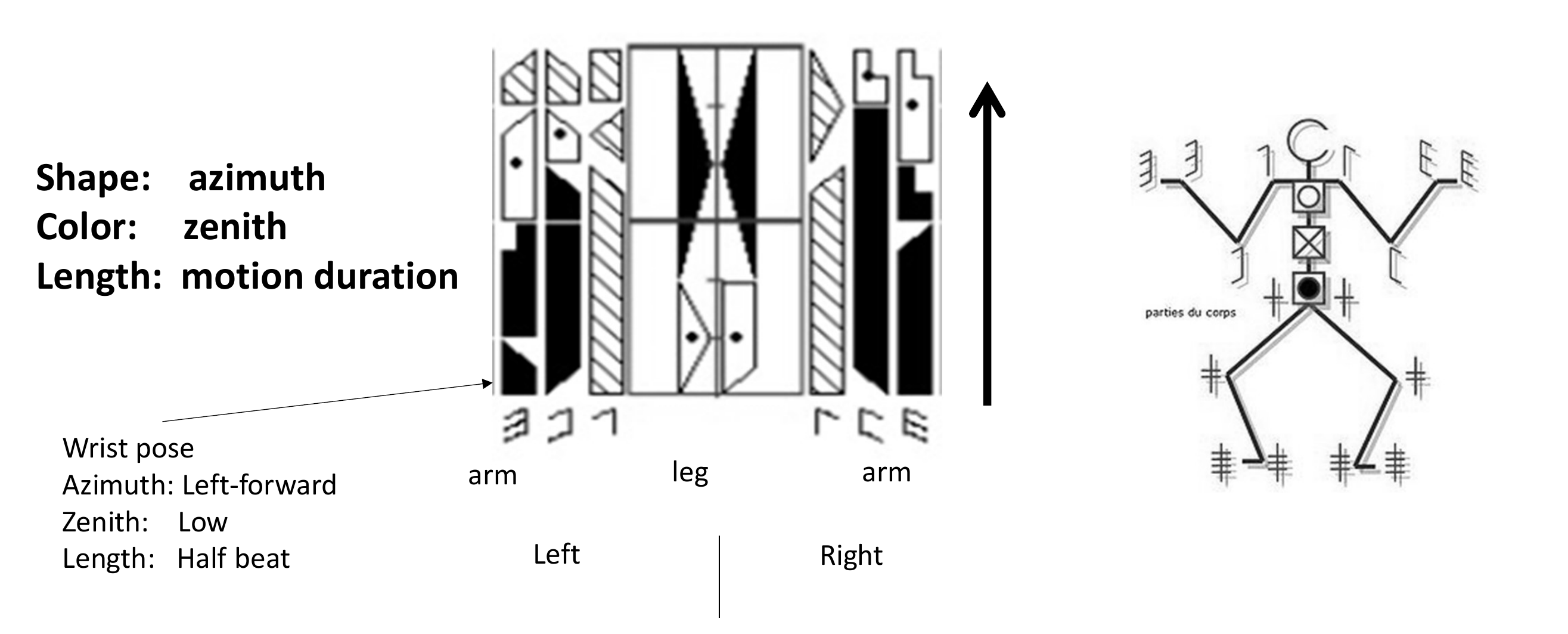}
    \caption{Labanotation}
    \label{fig:labanotation}
\end{figure}

Fig.~\ref{fig:labanotation} shows a Labanotation score. In a music score, time flows from left to right, whereas in a Labanotation score, time flows from bottom to top. When humans look at a series of movements, they do not see a continuous movement, but rather focus on postures at certain points in time (mainly at short stops), and, then, interpolate them to understand a continuous movement. Labanotation follows this digitization in the time domain, i.e., each symbol in the score represents a posture at each brief stop. Each column of the Labanotation is used to represent the postures of one human part, such as an arm or an elbow. The length of each symbol represents the time it takes to move from the previous posture to that posture. The shorter the symbol, the faster the person moves the part; the longer the symbol, the slower the person moves the part. The shape of the symbol represents the digitized azimuth angles of the part in eight directions, such as east, west, south, north, and south, and the texture of the symbol represents the digitized zenith angles of the part in five directions, such as zenith, higher, level, lower, and nadir. Although the digitization of the eight azimuthal angles and the five zenithal angles seems somewhat coarse, it is consistent with Miller's theory of human memory capacity~\cite{miller1956magical}, which is probably why the dance community has represented angles with this granularity.

LabanSuite~\cite{IkeuchiIJCV2018} was developed to detect short pauses, digitize the postures at the pauses, and obtain a Labanotation score from a human movement sequence. There are 2D and 3D versions: the 3D version uses a bone tracker from kinetic input, while the 2D version uses a bone tracker based on OpenPose~\cite{cao2017realtime} and lifting~\cite{rayat2017exploiting} from video input. Both versions still extract short stops from the bone motion sequence and digitize postures at the graunlarity of 8 azimuthal and 5 zenithal angles according to the Labanotation rule.

\begin{figure}[h]
    \centering
    \includegraphics[height=45mm]{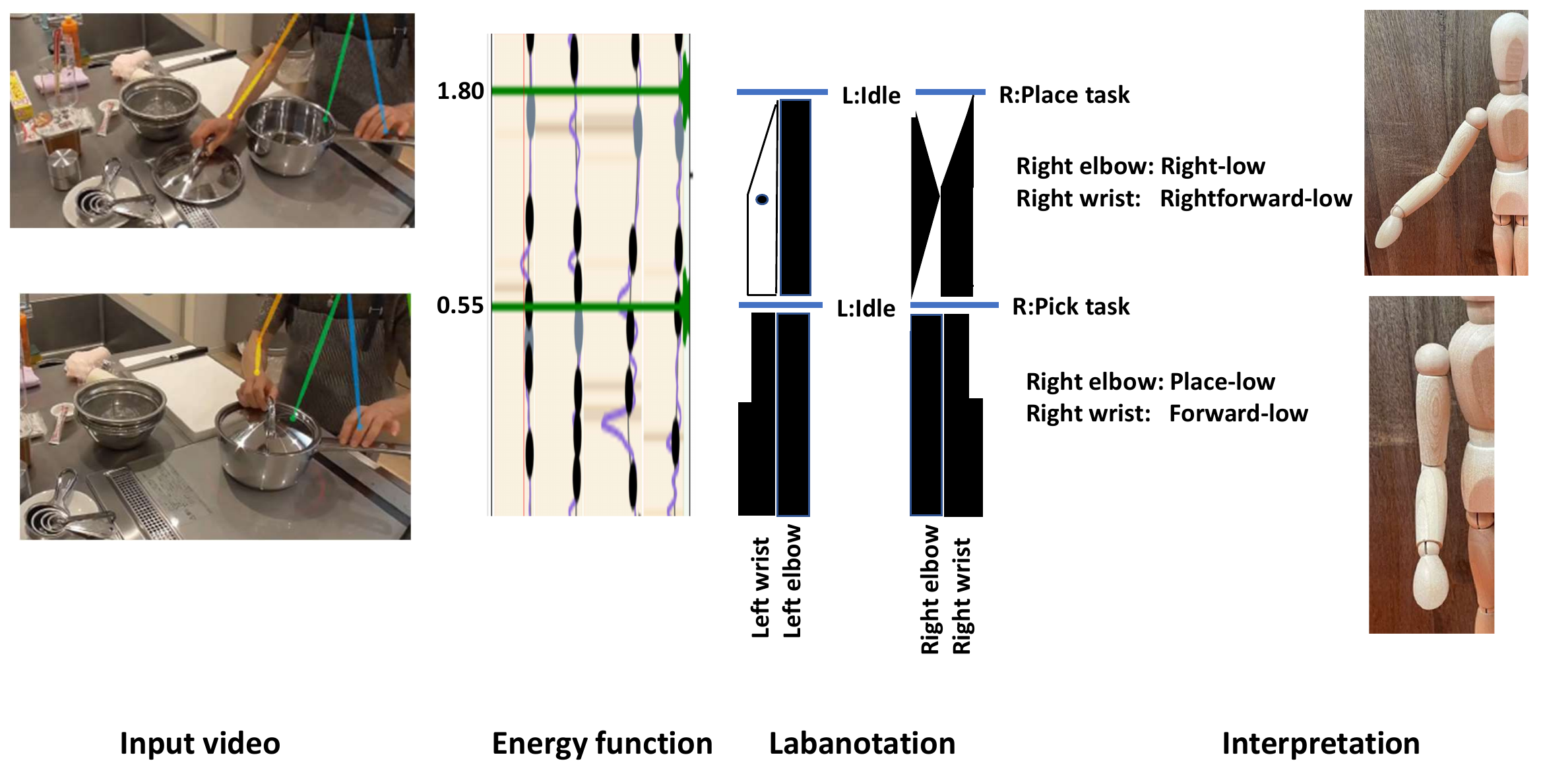}
    \caption{Applying LabanSuite to a cooking scene.}
    \label{fig:cooking}
\end{figure}

Fig.~\ref{fig:cooking} shows the 2D version applied to a cooking scene. Brief pauses are extracted at the 0.55 and 1.80 seconds and the postures at these timings are described as Labanotation symbols by the LabanSuite. These correspond to the times when the right hand picks the lid of the pod (0.55) and places it on the table (1.80). The right hand at the pick timing is recorded as ``Place low"  for the right elbow and ``Forward low" for the right wrist, and the right hand at the place timing is recorded as ``Right low" for the right elbow and ``Right forward low" for the right wrist. These postures are recorded in the task models and used for robot execution.

\subsection{Body-tool relation}
\begin{figure}[h]
\begin{tabular}{ccc}

\begin{minipage}{0.3\hsize}
\centering
\includegraphics[height=20mm]{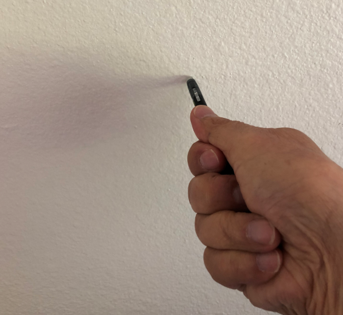} \\
(a) Power
\end{minipage}

&

\begin{minipage}{0.3\hsize}
\centering
\includegraphics[height=20mm]{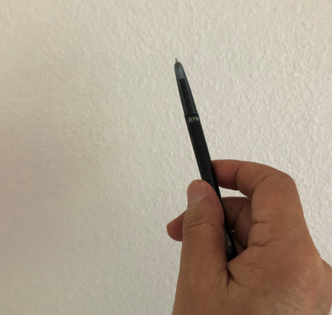} \\ 
(b) Control
\end{minipage}

& 

\begin{minipage}{0.3\hsize}
\centering
\includegraphics[height=20mm]{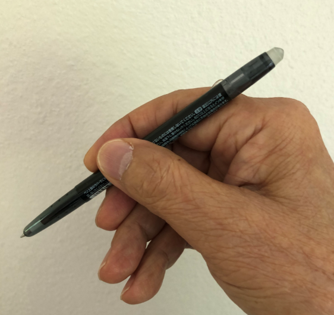} \\ 
(c) Versatility
\end{minipage}
\end{tabular}

\caption{The grasping method depends on the purpose of the task sequence. (a) for applying power. (b) for precise control. (c) for versatility}
\label{fig:grasp}

\end{figure}

The relationship between the body and the tool when performing a task sequence, especially how the tool is grasped, is an important factor in the success of the task sequence. As shown in Fig.~\ref{fig:grasp}, even when we grasp the same pen, 
we grasp it differently depending on the purpose of the task sequence. For example, when pushing, we grab the pen with the whole hand so that we can apply enough force to the pen (Fig.~\ref{fig:grasp}(a)), and when pointing, we pick the pen with the fingertips so that we can freely manipulate the direction of the pen (Fig.~\ref{fig:grasp} (b)). When writing, we hold it so that we can control the tip of the pen while exerting pressure on it, as shown in Fig.~\ref{fig:grasp} (c). 

\begin{figure}[h]
    \centering
    \includegraphics[height=30mm]{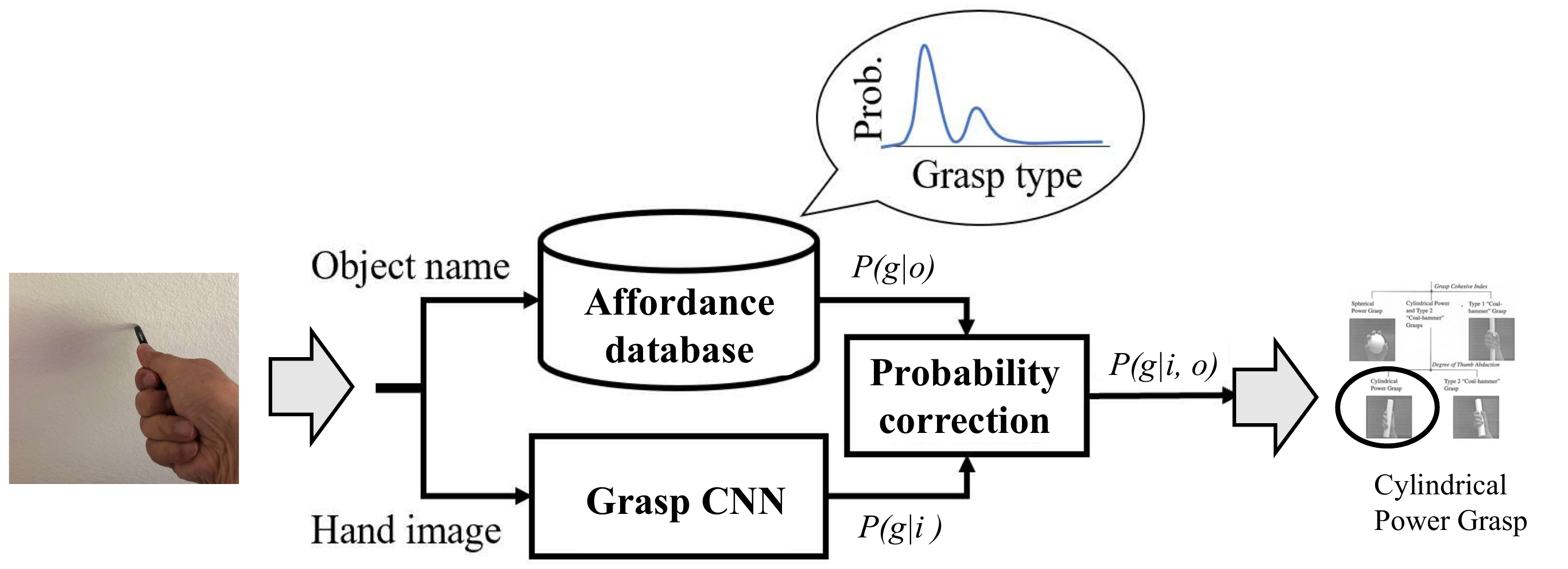}
    \caption{Grasp recognition system. The system is based on CNN and enhanced by grasp afforance of candidate objects.}
    \label{fig:wake-grasp}
\end{figure}

Various grasping taxonomies have been proposed in the robotics community, starting with Cutkosky's pioneering work~\cite{cutkosky1989grasp} to Felix's recent detailed taxonomy~\cite{feix2015grasp}. For LfO, the concept of the contact web~\cite{kang1997toward} was used to create a grasp taxonomy. Consistent with the application of the closure theory (discussed later), we use this contact web-based taxonomy and built a recognition system based on it, (a system that classifies an input image into a grasp type in the taxonomy using a CNN). To improve the performance, we used the prior distribution of grasp types (affordance) for each object~\cite{wake2020grasp}. See. Fig.~\ref{fig:wake-grasp}.

\subsection{Tool-Environmental relation}
\begin{figure}[h]
    \centering
    \includegraphics[height=30mm]{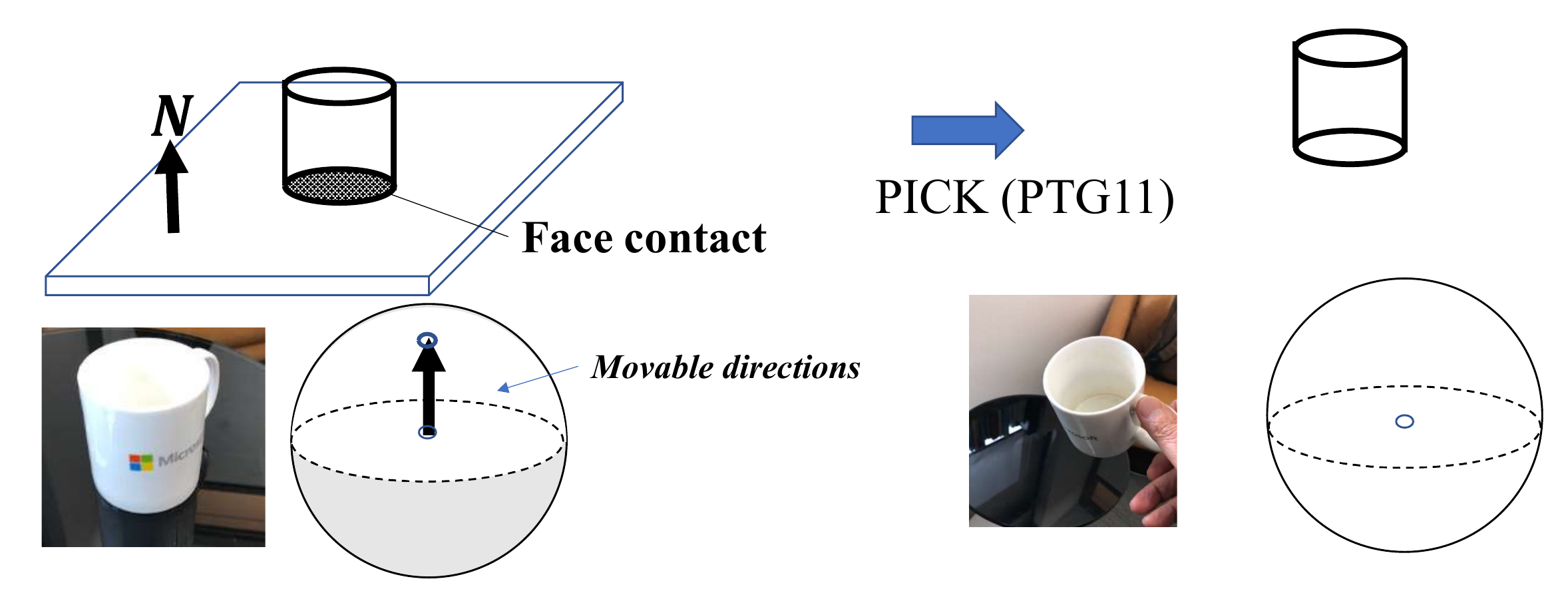}
    \caption{Pick task}
    \label{fig:pick}
\end{figure}

We define a task as a transition in the contact relation between the grasped object and the environment as in Ikeuchi et. al.~\cite{ikeuchi1994toward}. As an example of a state transition, let us consider a pick task. In Fig.~\ref{fig:pick}, the cup on the table is in surface contact with the table surface before the pick task. Surface contact constrains the range of directions an object can move. Let $N$ be the normal direction of the tabletop surface and $X$ be the direction in which the cup can move:
\begin{equation}
N \cdot X \geq 0
\end{equation}
The only directions in which the cup can move are up in Fig.~\ref{fig:pick} are upward, i.e., the northern hemisphere of the Gaussian sphere, assuming the table surface is facing upward. The Gaussian sphere is used to illustrate the direction of movement; a unit directional vector is represented as a point on the Gaussian sphere. The starting point of the vector is located at the center of the sphere, and the end point on the sphere represents the directional vector. In the example of a cup on a table, the upper directions are the movable directions of the cup, which correspond to the northern hemisphere of the Gaussian sphere shown in white, assuming the surface normal of the table is represented as the north pole of the Gaussian sphere.

As a result of the pick task, the cup is lifted into the air and no longer has the surface contact with the table. As the result, the cup can move in all directions. The entire spherical surface becomes the movable region. This transition from one-directional contact to no contact is defined as the pick task. See Fig.\ref{fig:pick}.

Let us enumerate the surface contact relations to count up how many transitions we have between them in general. For each additional surface contact, the range of motion is further constrained by an additional linear inequality equation corresponding to the surface contact. Adding an inequality equation one by one, the final range over which the object can move can be expressed as the solution of linear simultaneous inequalities given by the set of surface contacts. Using Kuhn-Tucker theory with respect to the solution space of linear simultaneous inequalities, we can count seven characteristic solution spaces, or contact states, for infinitesimal translational motion and seven characteristic solution spaces, or contact states, for infinitesimal rotational motion. Since there are seven possible initial states and seven possible final states, in principle, we have 7 x 7 = 49 transitions for  translation and 49 transitions for rotation. However, by considering the physical constraints, the number becomes 13 transitions for translation and 14 transitions for rotation~\cite{ikeuchi2021semantic}. Thus, the maximum number of tasks required would be 27 different tasks, assuming all possible transitions.

Further examinations in household tasks using Youtube video as well as our own-recorded cooking video reveals that only a few of them actually occur:
\begin{itemize}
\item{\bf PTG1:} A Pick (PTG11) infinitesimal translation causes a transition from semi-DOF (surface contact) to the full DOF, and its reverse movement by Place (PTG13).
\item{\bf PTG3:}  A Drawer-open (PTG31) infinitesimal translation causes a transition that semi-DOF with two directional constraint degrees is lost and results in a two-directional constraint degrees, and its reverse movement by Drawer-closing (PTG33).
\item{\bf PTG5:} A Door-opening (PTG51) infinitesimal rotation causes a transition that semi-DOF with two-directional constraint degrees is lost and results in a two-directional constraint degrees and its reverse movement by Door-closing (PTG53).
\end{itemize}
Physical transitions other than these rarely occur in household activities, probably because they are difficult for the average unskilled person to perform. In this paper, these six types of tasks have been prepared as the task models in the manipulation skill library. Note that the remaining tasks can be added to the library, if needed in the future.

In household tasks, in addition to the physical tasks described above, we need to consider semantic tasks. For example, consider an action such as wiping a table surface with a sponge. Physically, the sponge on the table surface, is in contact with the surface in only one direction. This means that the sponge can be moved upward beside moving on the table surface. However, if the sponge is lifted off the surface of the table, it cannot wipe the surface of the table. In other words, under the common sense of wiping, the movement must be such that it always maintains contact with the surface. To express this, we can introduce a virtual surface that exists parallel to the surface of the table. The wiping motion can be described by considering that the sponge can only move in directions between this virtual surface and the original physical table surface. We will refer this virtual surface as a semantic constraint surface. By examining the actions in the household operations, the following five semantic constraints were obtained.

\begin{itemize}
    \item{\bf SGT1: Semantic ping} A task such as carrying a glass of juice can be described by a semantic ping. Physically, the glass can rotate about any axis in the air. However, in order to carry the juice without spilling it, the glass is only allowed to rotate about the axis along the direction of the gravity and not in any other axis direction. We assume that the semantic ping stands perpendicular to the surface of the glass. The object is only allowed to rotate around the semantic ping.

    \item{\bf STG1: Semantic wall} In a task such as wiping a table, only the directions of motion along the surface of the table, i.e., between the actual table surface and the semantic surface, is allowed.

    \item{\bf STG3: Semantic tube} Tasks such as peeling in cooking require translational motion along a specific trajectory on the object surface. This motion can be interpreted as the one along a semantic tube.

    \item{\bf STG4: Semantic sphere} In the STG2 semantic wall, motion was constrained between two planes. In the case of wiping spherical surfaces, motion is constrained between two spherical surfaces, a real sphere and a semantic sphere.
   
    \item{\bf STG5: Semantic hing} In a task such as pouring water from one pitcher into another, the motion of the pitcher must be a rotational motion around the spout, meaning rotation around the semantic hinge at the spout.
    \end{itemize}

\begin{figure}[h]
    \centering
    \includegraphics[height=50mm]{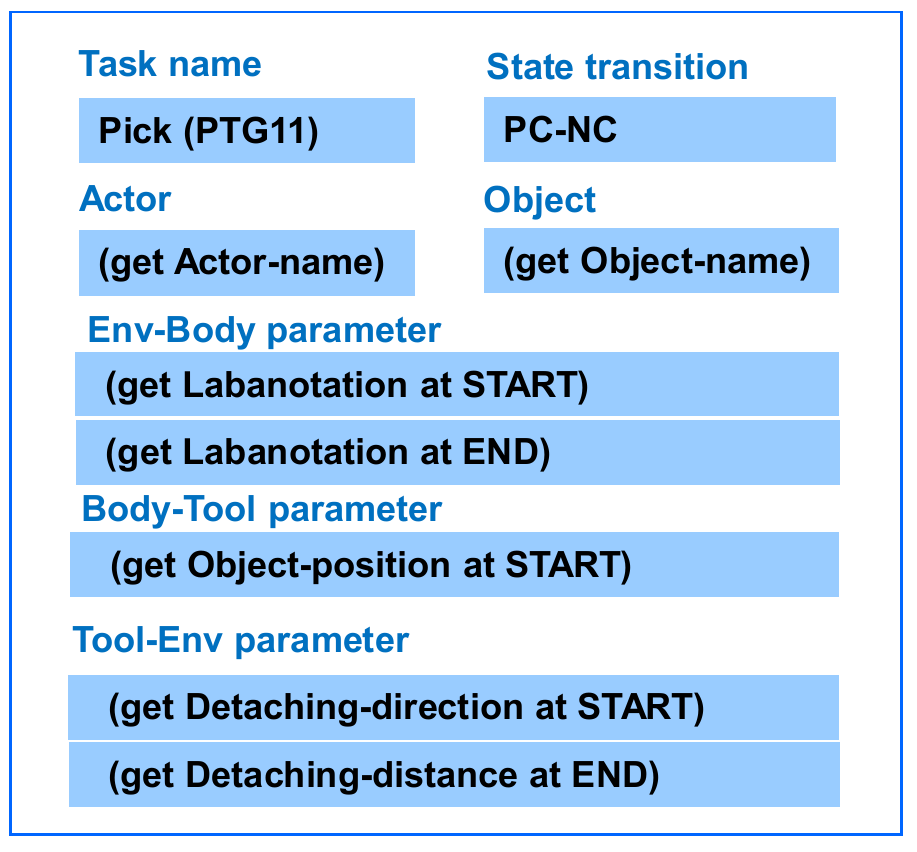}
    \caption{An illustrative example of Pick (PTG11) task model}
    \label{fig:taskmodel}
\end{figure}

Task models are designed in the design phase, with the name and slots for the skill parameters essential for task execution, as in a Minsky's frame. Fig~\ref{fig:taskmodel} shows Pick (PTG11) as an example task model. The task name is registered as Pick (PTG11), and the transition of surface contacts is described as PC (partial contact), i.e., one-directional contact, to NC (no contact). This transition information will not be used for execution, but is provided for debugging purposes. The actor slot describes whether to execute with the right or left hand. A slot to register the name of the object is also prepared, which is used for MS-Custom vision to identify the location of the object in the image. Slots are also prepared for the first and last Labanotation, the initial position of the object, and the detaching direction and distance, in which direction and how much the object will be lifted up. Daemons are attached to these slots to observe and obtain these parameters from the demonstration. 

During the teaching mode, the system, the task encoder, first recognizes which task it is from the verbal input and instantiates the corresponding task model. At the same time, it also obtains the name of the target object from the verbal input. Next, the instantiated task model collects the necessary skill parameters from human demonstration according to each demons attached at the slots in the task model.

\section{Skill Library}

In order for a robot to be able to perform the corresponding motion from the recognized task model, an execution module corresponding to each task model is necessary. In this section, we design the grasp skill library and the manipulation skill library which contain such execution modules, which we refer to as skills. The core of each skill consists of a control agent, Bonsai brain, with a policy fine-tuned using Bonsai reinforcement learning system. Each skill also includes interfaces to retrieve necessary skill parameters from the corresponding task model as well as to obtain additional parameters needed at run time.

\subsection{Grasp skill library}
\paragraph{Closure theory and contact web}

\begin{figure}[h]

\begin{tabular}{ccc}

\begin{minipage}{0.5\hsize}
\centering
\includegraphics[height=20mm]{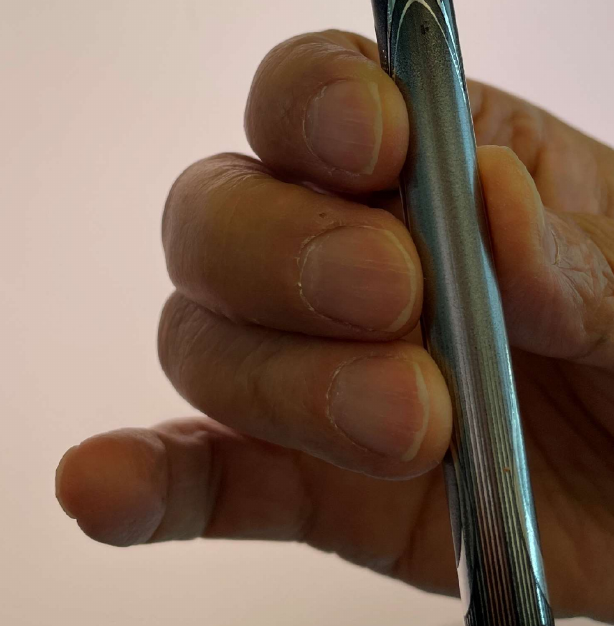} \\
(a) Thumb-3 grasping
\end{minipage}

&

\begin{minipage}{0.5\hsize}
\centering
\includegraphics[height=20mm]{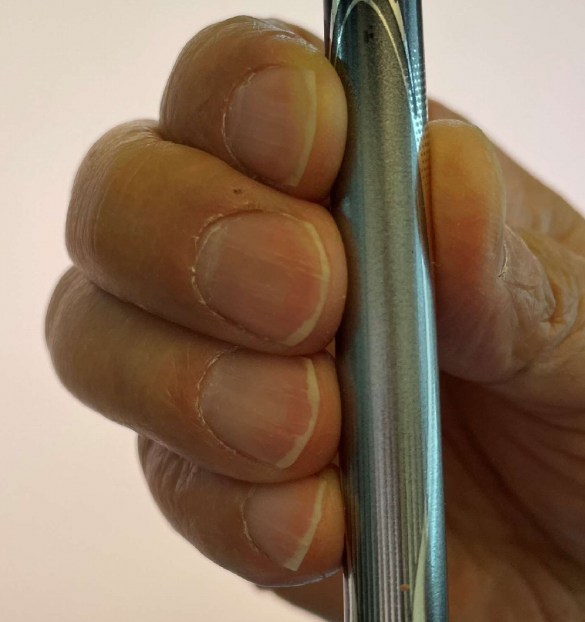} \\ 
(b) Thumb-4 grasping
\end{minipage}

\end{tabular}

\caption{Thumb-3 and Thumb-4 grasping. The grasp classfication contains many classfications that are redundant for the purpose of robot tasks. There is little difference between these two types of grasping in terms of supporting the pen.}
\label{fig:thumb}
\end{figure}

In terms of robot execution of task sequences, the 43 grasp recognition classifications are redundant and are aggregated using the closure theory. For example, Thumb-4 finger and Thumb-3 finger can achieve almost the same goal in the task sequence following the grasp. See Fig.~\ref{fig:thumb}. According to the closure theory, the grasping task can be classified into the following three objectives~\cite{yoshikawa1999passive}.

\begin{figure}[h]

\begin{tabular}{ccc}

\begin{minipage}{0.3\hsize}
\centering
\includegraphics[height=30mm]{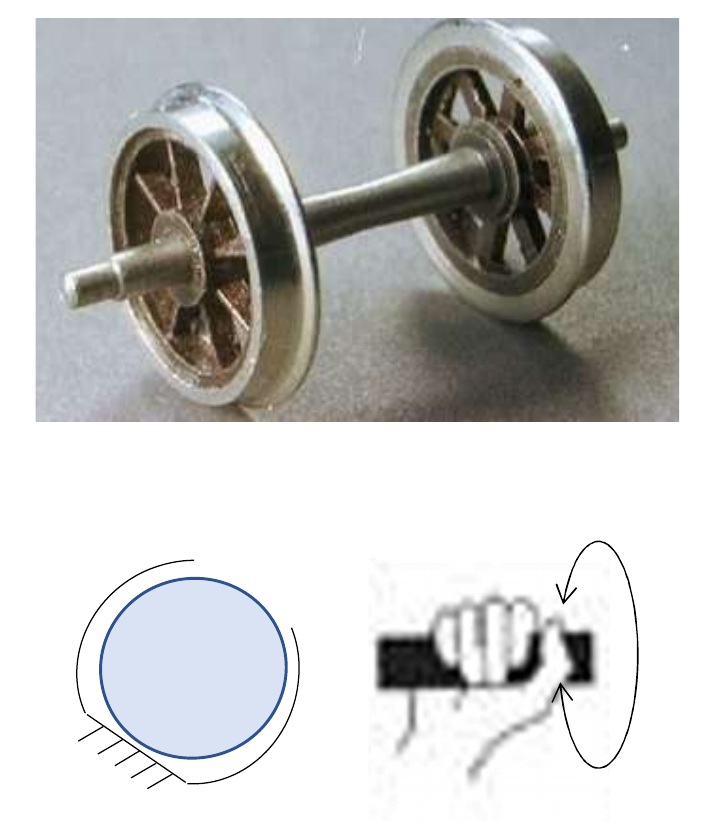} \\
(a) Passiveform closure
\end{minipage}

&

\begin{minipage}{0.3\hsize}
\centering
\includegraphics[height=30mm]{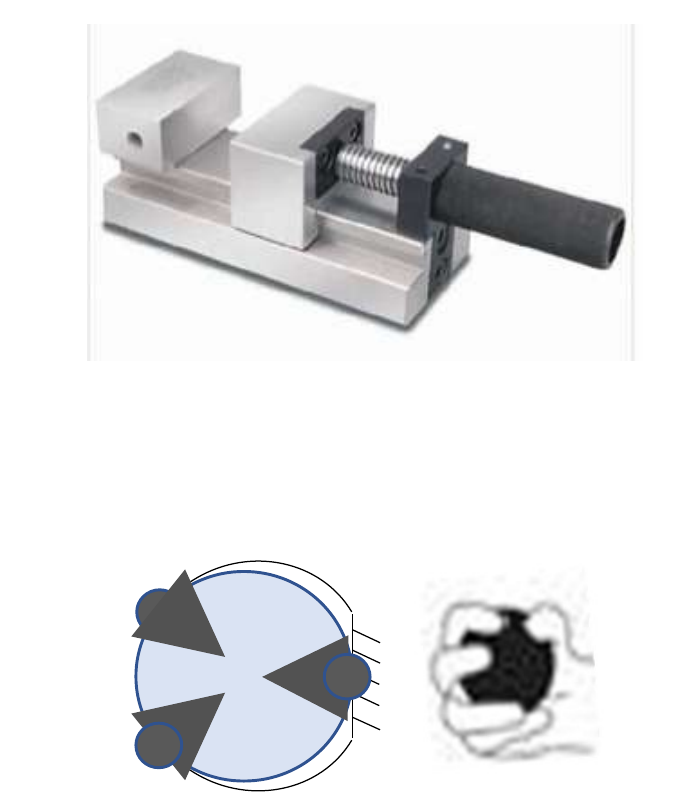} \\ 
(b) Passiveforce closure
\end{minipage}

&

\begin{minipage}{0.3\hsize}
\centering
\includegraphics[height=30mm]{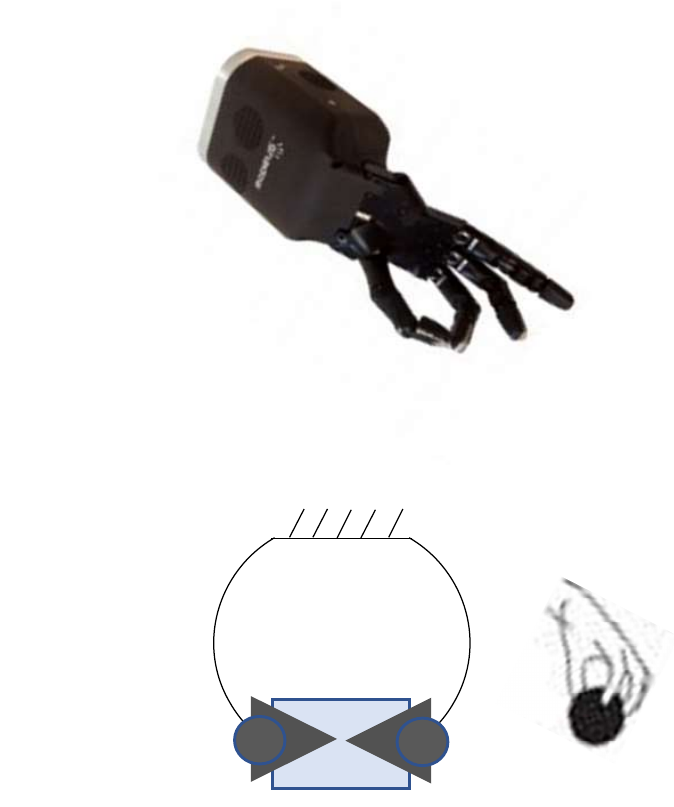} \\ 
(c) Activeforce closure
\end{minipage}
\end{tabular}

\caption{Three closures}
\label{fig:closure}
\end{figure}

\begin{itemize}
    \item{\bf Passive form closure:} to maintain the constant position of a grasped object by bringing the hand into a particular shape without actively applying force on the grasped object. Examples include wheel bearing in a wheel. See Fig.~\ref{fig:closure}
    \item{\bf Passive force closure:} to hold a grasped object in place without loosening its motion by applying force from all directions. Examples include a vice.
    \item{\bf Active force closure:} to allow fingertip manipulation of the grasped object while grasping it.
\end{itemize}

To be able to perform these three types of closures, we assume a particular gripper and a particular shaped object, and computer contact-webs, the distribution of contact points between the object and the gripper. In this paper, Shady-hand is used as the gripper and the shape of an object is assumed to be able to approximate as a superquadric surface. Three kinds of contact-webs are computed for each object shape: a passive-form contact web, passive-force contact web, and an active-force contact web. 

\begin{figure}
    \centering
    \includegraphics[height=30mm]{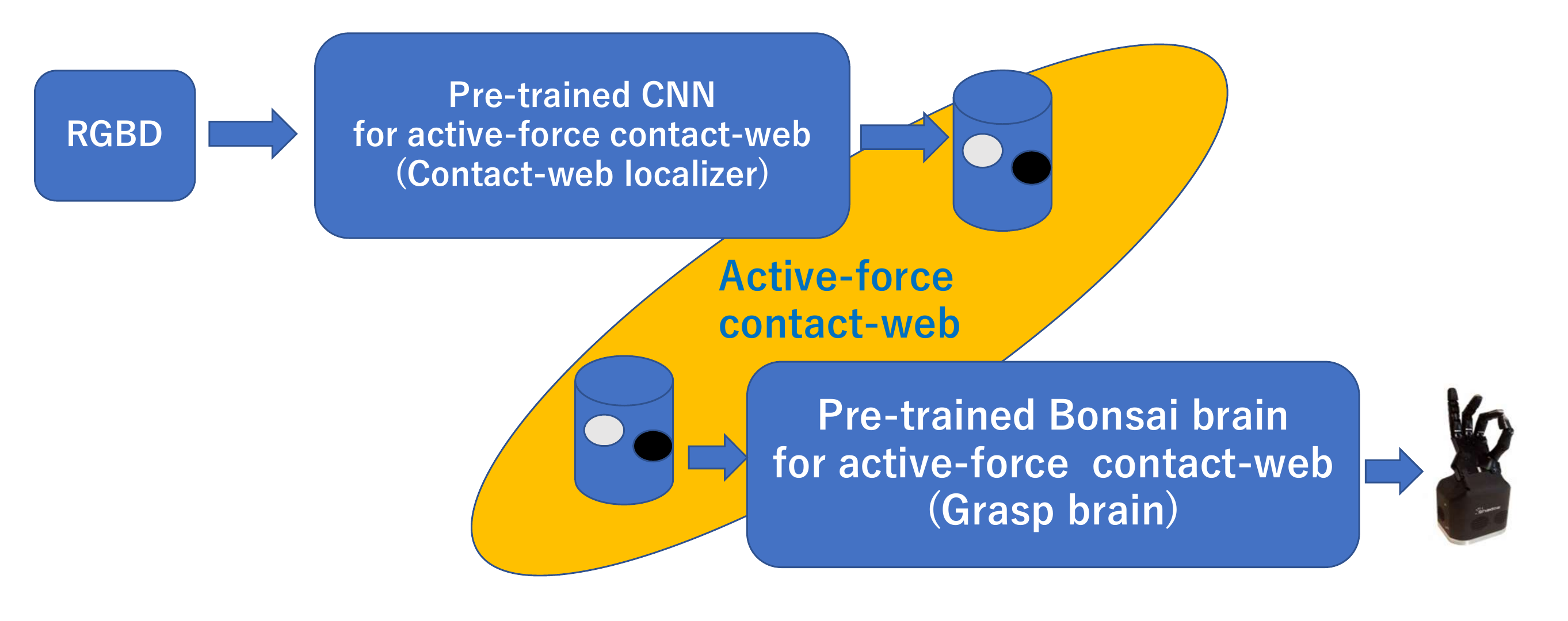}
    \caption{A grasp-pipe line. The observation module and the execution brain are designed as an integrated unit. The observation module consists of a CNN that determines the positions of the contact web from the scene, and the execution brain is trained using reinforcement learning to control the hand based on the positions of the contact web and the force feedback.}
    \label{fig:grasp-pipe}
\end{figure}

In order for a robot to perform a grasping task, the skill must obtain the location of the contact webs from the visual information at run time to guide.
As shown in Fig.~\ref{fig:grasp-pipe}, the visual and grasping modules are designed as an integral part in the grasping skill. The visual CNN determines the contact-web on the target object, and then the grasping brain relies on this contact-web location to control the hand. Such pipelines are prepared for each of the three contact-webs. At runtime, a specific grasp pipeline is activated by a grasp task model. Note that only three pipelines are prepared that are independent of object shape and size, since differences in shape and size are absorbed by the randomization in CNN training and reinforcement learning.

\paragraph{Contact-web localizer}

The contact-web localizer consists of a multi-layer CNN that takes a depth image as input and outputs the contact-web locations.
In order to obtain contact web locations without making a priori assumptions about the shape and size of an object, the depth image training data was generated by randomly sampling the size, shape, and viewing parameters of a superquadric surface, which can approximate a variety of objects by varying the shape and size parameters. Specifically, the shape parameters, the size parameters, the azimuth angle, and the zenith angle are randomly selected from the range of 0 to 1.0, 10 cm to 30cm, between plus and minus 120 degrees from the front, and 0 to 90 degrees, respectively. For each closure, the positions of the contact-web were calculated analytically based on the superquadric equation of these parameters using the specific robot hand (in this paper, the Shady-hand). The axis directions and the origin of the coordinate system are also given as true values so that the object-centered coordinate system of the superquadric surface can also be output.

\paragraph{Grasp skills}
Each grasp skill contains a Bonsai brain, an agent with a policy for controlling the hand. The brain is pre-trained offline using the Bonsai reinforcement learning system to control the robot's hand movements. Note that for reusability at the hardware level, only the hand movements are trained, not the whole arm, to take advantage of the effectiveness of the role-division algorithm, which will be discussed later. The brain receives the positions of the contact-web from the contact-web localizer and the approach direction from the demonstration as hint information with respect to the object centered coordinates. The observable states include the current joint positions of the hand and the drag force derived from the effort at each joint.  Since it is not practical to attach a force sensor to each fingertip, we decided to measure the drag force on a fingertip using the effort value instead. The effort is defined as the amount of electrical current required by the joint motor to move the joint to the target position. The greater the reaction force, the more current is required. The grasp task is considered complete when the drag force on the fingertip, as measured by the effort, is sufficient. The reward is the success or failure of the pick after the grasp. 

\begin{figure}[h]
    \centering
    \includegraphics[height=40mm]{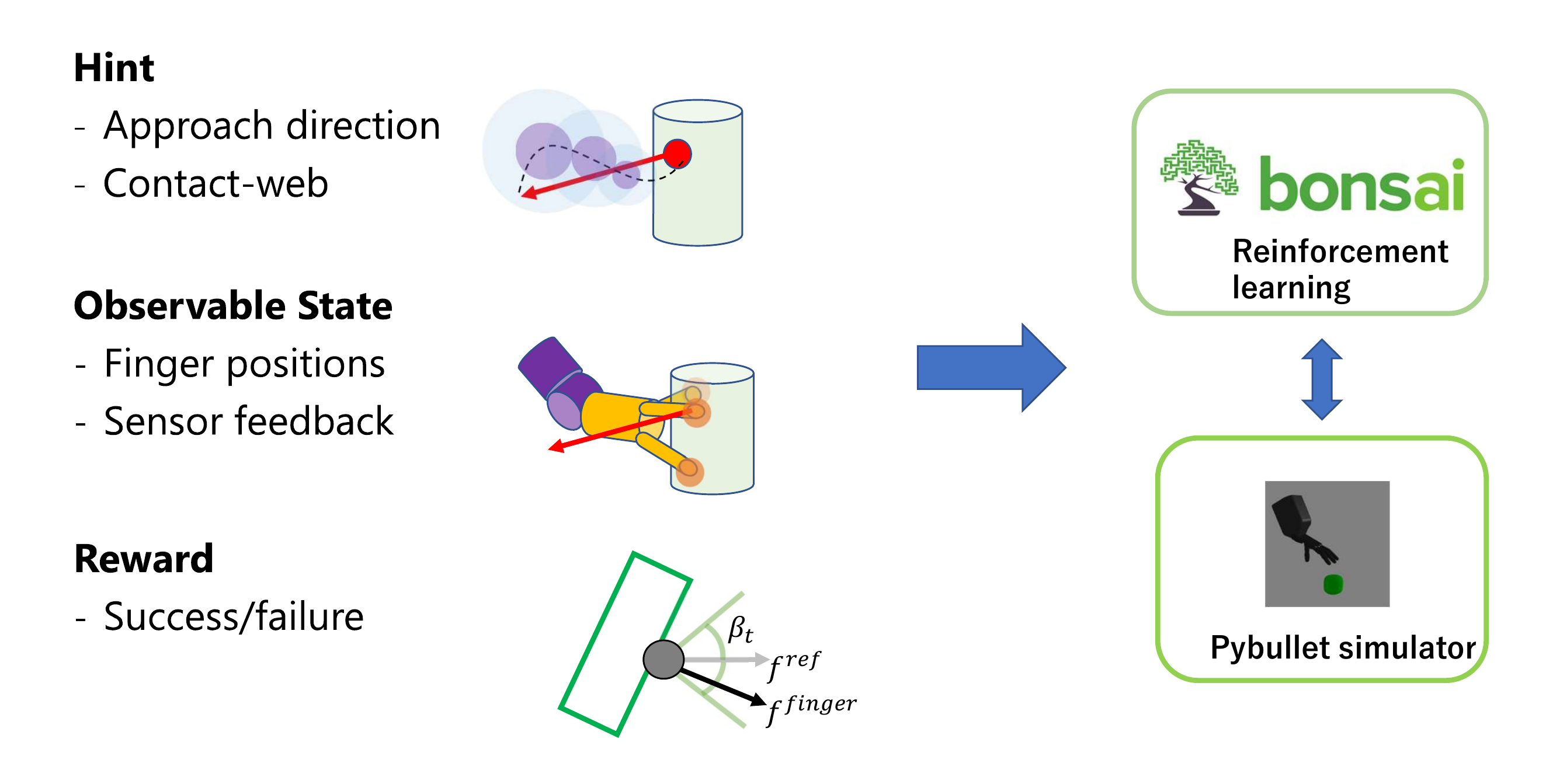}
    \caption{Framework of reinforcement learning}
    \label{fig:RL}
\end{figure}

Fig.~\ref{fig:RL} shows the framework of reinforcement learning.

\subsection{Manipulation skill library}
The manipulation skill library is a collection of skills for manipulating a grasped object. Theoretically, 27 different skills are necessary, but considering the household domain, we have chosen to prepare 6 physical skills and one semantic skill in this implementation:
\begin{itemize}
    \item Pick(PTG11)/Place(PTG13)
    \item Drawer-opening(PTG31)/Drawer-closing(PTG33)
    \item Door-opening(PTG51)/Door-closing(PTG53)
    \item Bring-carefully (STG12)    
\end{itemize}

These skills are also implemented as brains using the Bonsai reinforcement learning system. In the manipulation skills, objects can be assumed to be already grasped. The hint information is the direction of motion of the grasped object by human demonstration. The observation states include the current joint positions and the drag force from the environment of the grasped object. The drag force is assumed to be obtained from a force sensor attached to the arm. The tasks of Place (PTG13), Drawer-closing (PTG33), and Door-closing (PTG53) are terminated when drag force from the wall or table surface is detected. The tasks of Pick (STG11), Door-opening (PTG31), and Drawer-opening (PTG51) ends when drag force is eliminated and the hand position given in the demonstration is reached. 
The reward for PTG13 is given by whether the object is stable when released, while the reward for the remaining tasks is given by whether the terminal condition is met.

\section{Implementation}
The system consists of two main modules:
\begin{itemize}
    \item {\bf Task encoder:} Recognizes tasks from verbal input, obtains skill parameters required for each task from visual input, and completes a sequence of task models.
    \item {\bf Task decoder:} Based on the task model sequence, pick the corresponding skills from the libraries and generates robot motions by the skills.
\end{itemize}

\subsection{Task encoder}

\begin{figure}
    \centering
    \includegraphics[height=40mm]{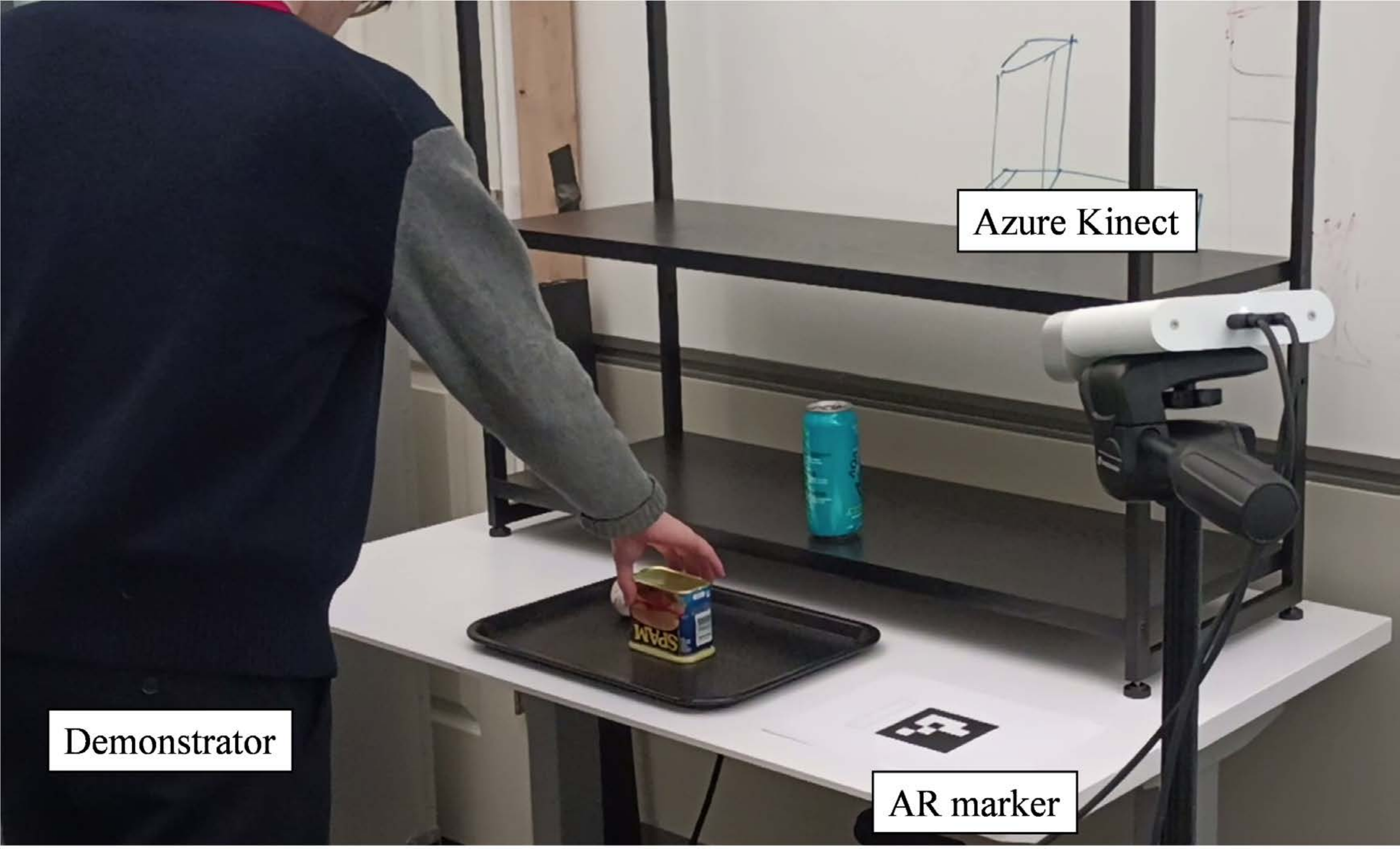}
    \caption{Human demonstration site}
    \label{fig:demo_site}
\end{figure}

Figure~\ref{fig:demo_site} shows the overview of the demonstration site for task encoding. An Azure Kinect sensor~\cite{k4a} was positioned to provide a full view of the site through the entire demo, capturing RGBD images of the demonstrator and the object as well as the audio input of the demonstrator. The resolution of the images was 1280x720 and the nominal sampling rates of the video and audio were 30 Hz and 48000 Hz, respectively. AR markers were placed to align the orientation of the demonstration coordinates with the robot coordinates. Note that the role of the AR markers is to align the rough orientation of both coordinates, and the objects, human and robot positions do not need to be exactly the same during demonstration and execution. 

One unit of robot teaching begins with the grasping task of an object, followed by several manipulation tasks of the object, and ends with the releasing task of the object. This sequence of grasping, manipulation, and releasing tasks is referred to as a GMR operation, and each GMR operation is assumed to manipulate the same object with the same hand. 

To facilitate teaching the granularity of tasks in one GMR operation, a stop-and-go method is used~\cite{wake2022interactive}. That is, for each task, the user verbally indicates the task to be performed, and starts moving his/her hand. Then, when the demonstration of that task is completed, the user stops moving his/her hand and, after a brief stop, begins the next cycle of verbal instruction and visual demonstration corresponding to the next task. This series of verbal instructions and visual demonstrations is repeated until the GMR operation is completed. For example, the GMR operation "pick up a cup and carry it to the same table" consists of five cycles: grasp the cup (passive-force grasp), pick it up (PTG11), bring it carefullly (STG12), place it (PTG13), and release it (release). 

\begin{figure}
    \centering
    \includegraphics[width=\linewidth]{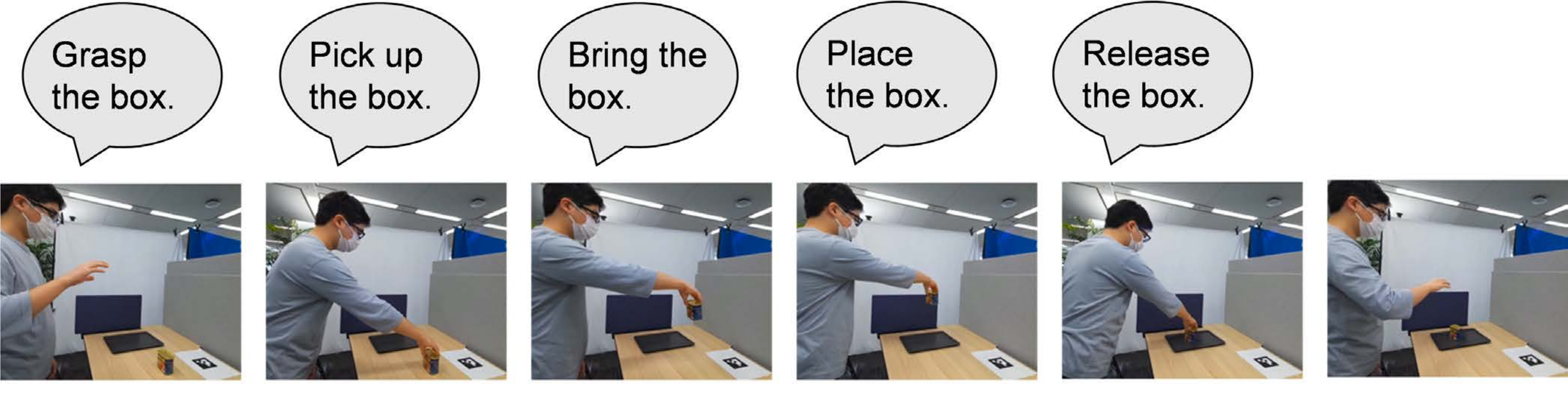}
    \caption{An example of human demonstration (Box-displacement demo)}
    \label{fig:demonstration}
\end{figure}

Figure~\ref{fig:demonstration} shows an example of human demonstration. The video and audio are segmented at the timings when the hand stops. To detect these timings from the input video, the brightness disturbance of the input video is characterized~\cite{ikeuchi1994toward}. For the calculation, the RGB image is converted to a YUV image and the Y channel is extracted as brightness. The brightness image is spatially filtered using a moving average of a 50 x 50 window, and the absolute pixel-by-pixel difference between adjacent frames is taken. The average of the differences is taken as the brightness perturbation at each timing. After removing outliers and low-pass filtering at 0.5 Hz, the local minima are extracted as the stop timings. The input video and audio are segmented based on these stop timings. This video-based segmentation is preferred over audio-based segmentation because, although human verbal instructions and hand demonstrations are roughly synchronized, their synchronization is not exact, and accurate video timings are needed for skill parameter extraction.

\begin{figure}
    \centering
    \includegraphics[height=40mm]{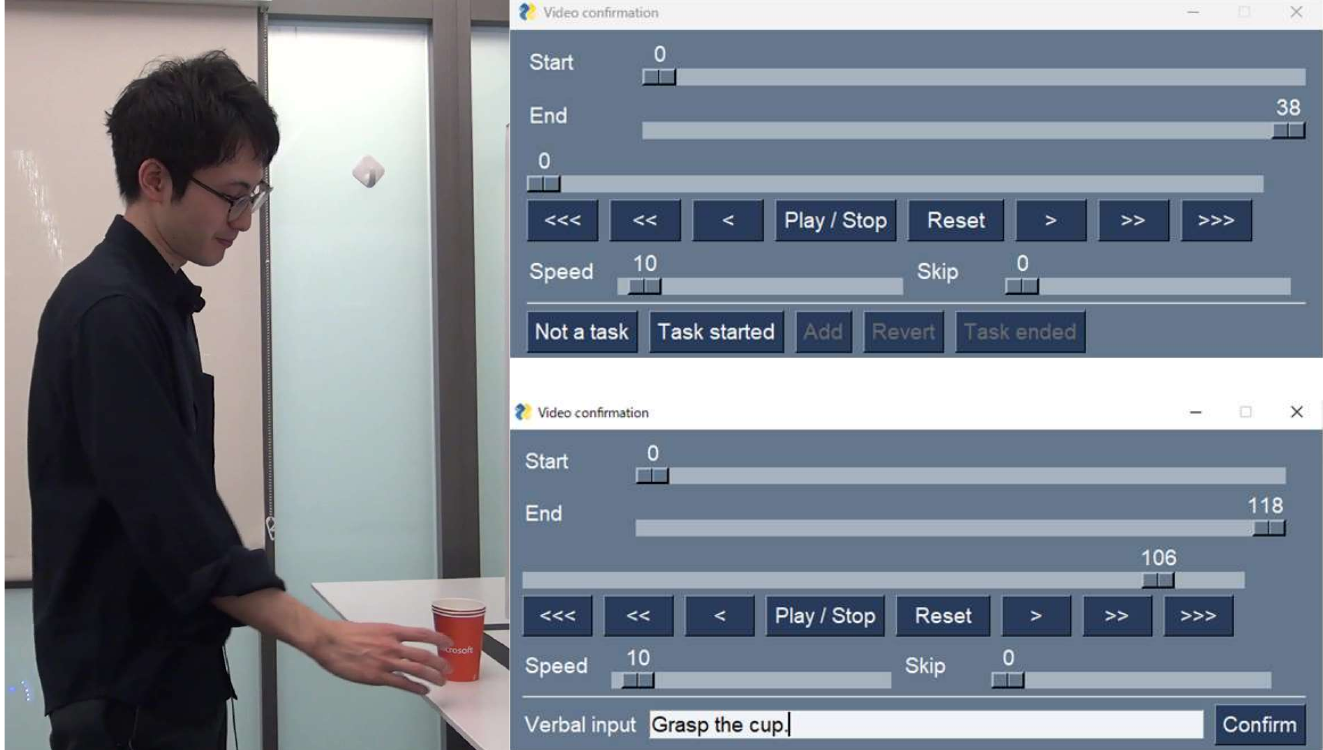}
    \caption{Preview GUI}
    \label{fig:GUI}
\end{figure}

The segmented video and audio are processed in two steps for task encoding. The first step is task recognition based on the audio segments, and the second step is skill parameter extraction based on the video segments. The segmented audio is recognized using a cloud-based speech recognition service~\cite{azure-speech-recognition}. In addition, the fluctuations in the user's verbal instructions are absorbed by a learning system based on our own crowdsourced data. The segmented audio segments, corresponding video segments, and recognition results can be previewed and modified by the user via the GUI, as shown in Fig.~\ref{fig:GUI}. 

\begin{figure}[h]
    \centering
    \includegraphics[height=40mm]{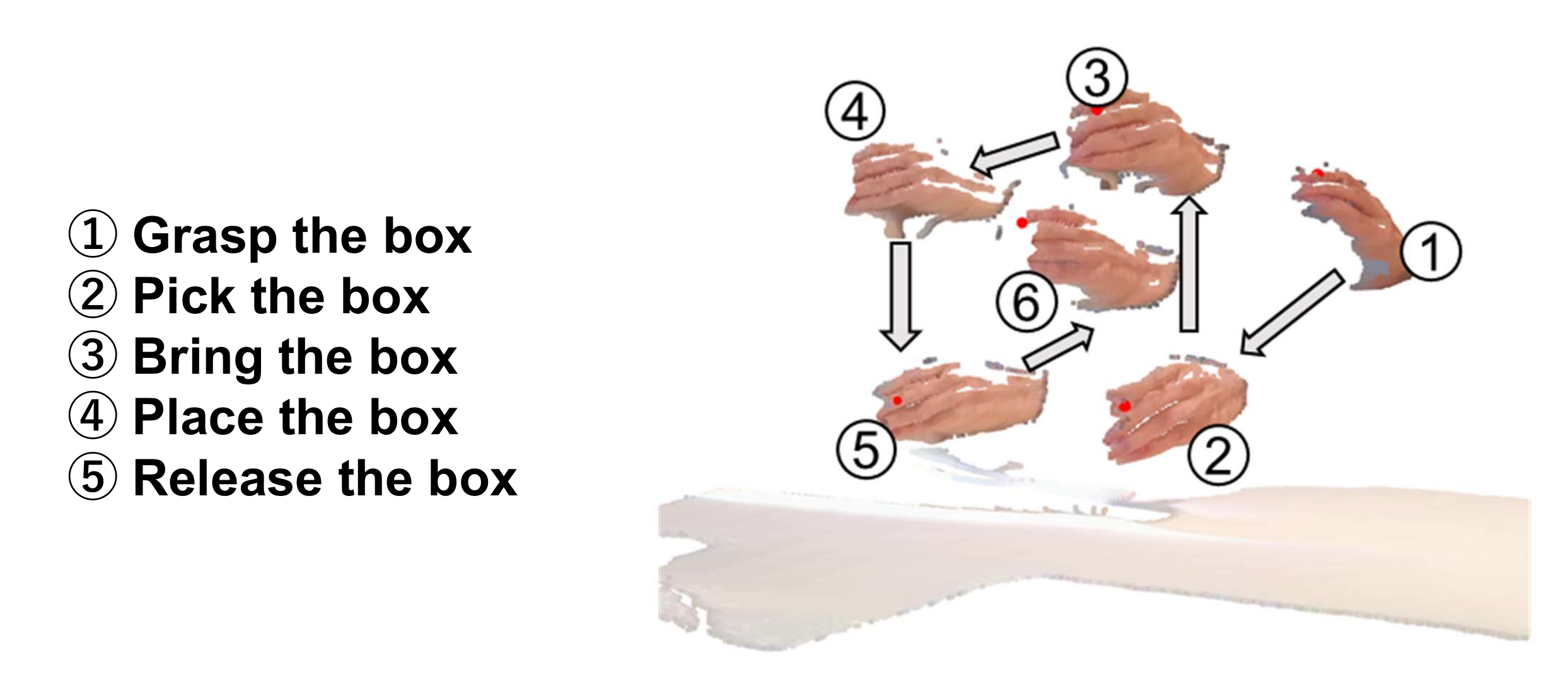}
    \caption{verbal instructions and recorded hand positions}
    \label{fig:hand-position}
\end{figure}

In the second step, the task-encoder instantiates a sequence of task models based on these recognition results. 
Each task model has a Minsky frame-like format with slots for storing skill parameters, which will be collected from the video segments. The skill parameters are mainly related to hand and body movements with respect to the target object. The hand movements are extracted using a 2D hand detector and depth images, as shown in Fig.~\ref{fig:hand-position}, while the body movements are obtained by the LabanSuite.

The image of the operating hand in the last frame of the grasping video is given to the grasp recognition system, which determines the grasp type. From the hand movements immediately before the end of a task, the approach direction for a grasping task, the attaching direction for PTG13 (place), PTG33 (drawer-closing), PTG53 (door-closing) are calculated in the object-centered coordinate system. The hand movements immediately after the start of a task also provide the departure direction for a releasing task and the detaching direction for PTG11 (pick), PTG13 (drawer-opening), PTG15 (door-opening). Human body movements are also important skill parameters. From the first and last postures of a human arm, we estimate the 3D posture of the performer and convert them into Labanotation using LabanSuite. These skill parameters are stored at each corresponding slot of a task model for later task decoding.

\subsection{Task decoder}

We developed the TSS system, a task decoding platform that allows a robot to execute task models given by the encoder~\cite{sasabuchi2023task}. The TSS system reads a set of task models from the task encoder, coordinates the parameters among the task models, calls the skills with Bonsai Brains corresponding to the task models, and passes the parameters stored in the task models to the skills. The TSS system can simulate movements with a virtual robot displayed in an unreal-engine environment or control a real robot via ROS. In addition, any skill in the skill sequence corresponding to a task sequence can be trained offline using the BONSAI reinforcement learning system.

\begin{figure}[h]
    \centering
    \includegraphics[height=60mm]{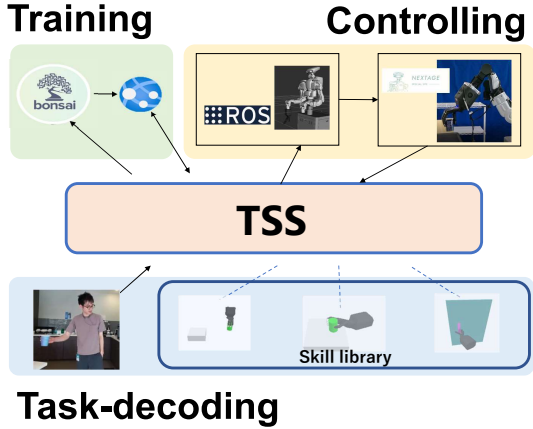}
    \caption{Task sequencing Simulator (TSS) for task decoding and skill refinement}
    \label{fig:TSS}
\end{figure}

The role-division algorithms in TSS improves reusability at the hardware level~\cite{SasabuchiRAL2021,takamatsu2022learning}. All skills in the libraries used by TSS are those trained to control robot hand movements, not the movements of the entire robot as a whole. To realize these hand movements, the robot arm must form certain poses given by the inverse kinematics (IK) of the robot. The shape of the arm, on the other hand, must be as close as possible to the shape of the human demonstration as represented by the Labanotation score; the adjustment between the arm shape given from the Labanotation score and the arm shape required by the Bonsai Brain is done using the Role-Division algorithm. Using this method, when introducing a new robot, only the Labanotation-based IK needs to be replaced, without need for retraining of skills.

\begin{figure}[h]
    \centering
    \includegraphics[height=40mm]{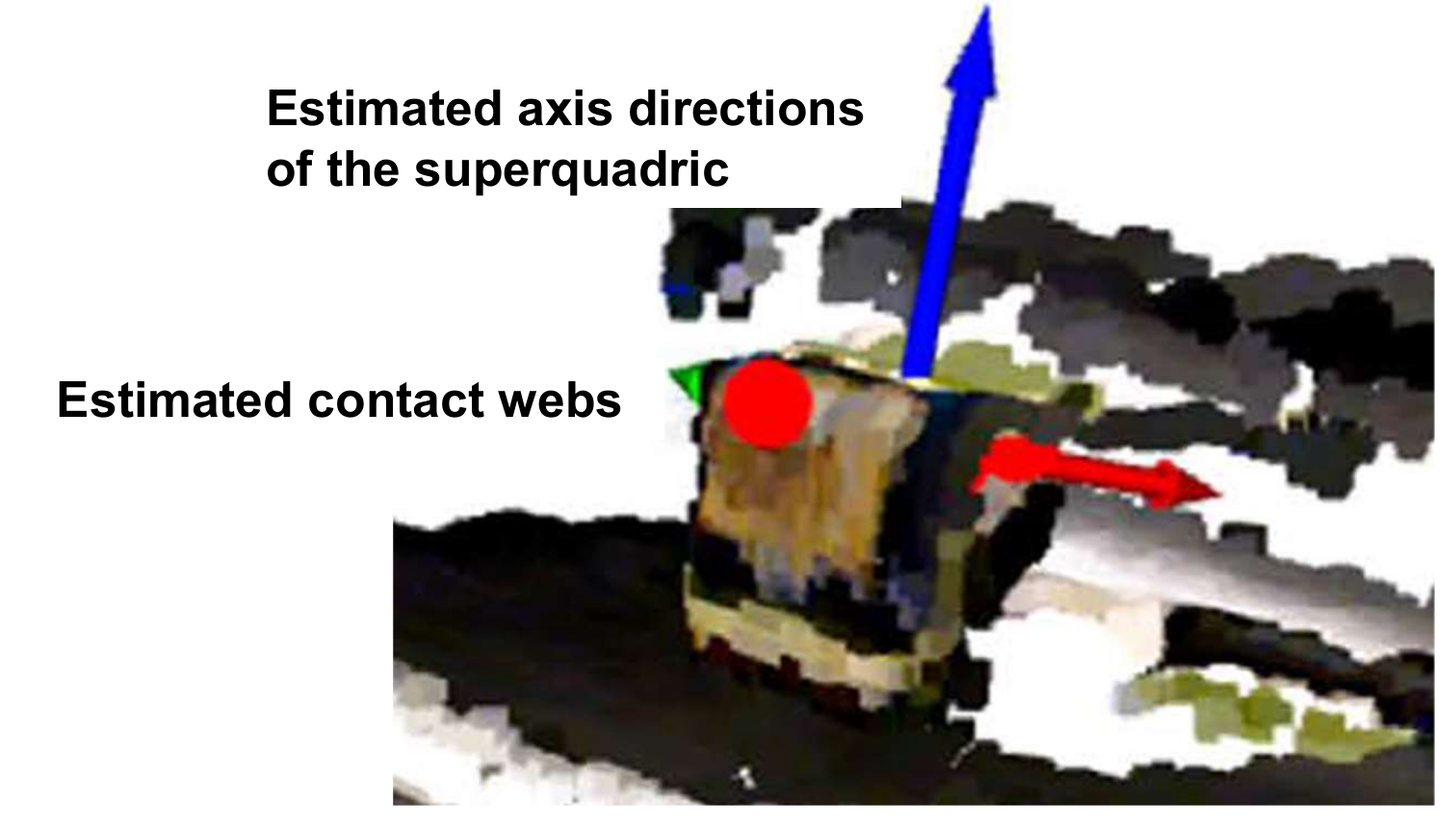}
    \caption{Estimated contact-web locations and axis directions of the superquadric. One point of the contact-web is located in front, depicted with a red dot. The remaining two points are behind and not shown in the figure. The CNN also estimates the local coordinate systems of the superquadric at the same time. These are shown by red and blue arrows.}
    \label{fig:depth}
\end{figure}

Three Bonsai brains in the three grasping skills are directly connected to their vision modules to determine the contact-web locations and the local coordinate system, as described in the grasp skill library section. This is because although the target objects are placed under nearly identical positions during the demonstration and execution, the robot's position is not exactly the same as the human demonstration position, so that the target object must be observed again during the execution in order to perform the robot grasp robustly.
From the grasping task model, the decoder activates the 3D sensor and detects the bounding box of the object using MS custom vision. Next, a plane fitting based on RANSAC (random sample consensus) [27] is applied to the depth image withing the bounding box so as to segment the object from the background table. The segmented depth image is then given to the trained CNN, which outputs the contact-web locations as well at the local coordinate system of the superquadric, from which the approach direction is re-calculated. Fig.~\ref{fig:depth} shows an example of the parameters obtained at execution time. Finally, the Bonsai brain receives those locations and the direction as  hint information, and generates hand and finger movements based on force feedback. 

\begin{figure}[h]
    \centering
    \includegraphics[height=40mm]{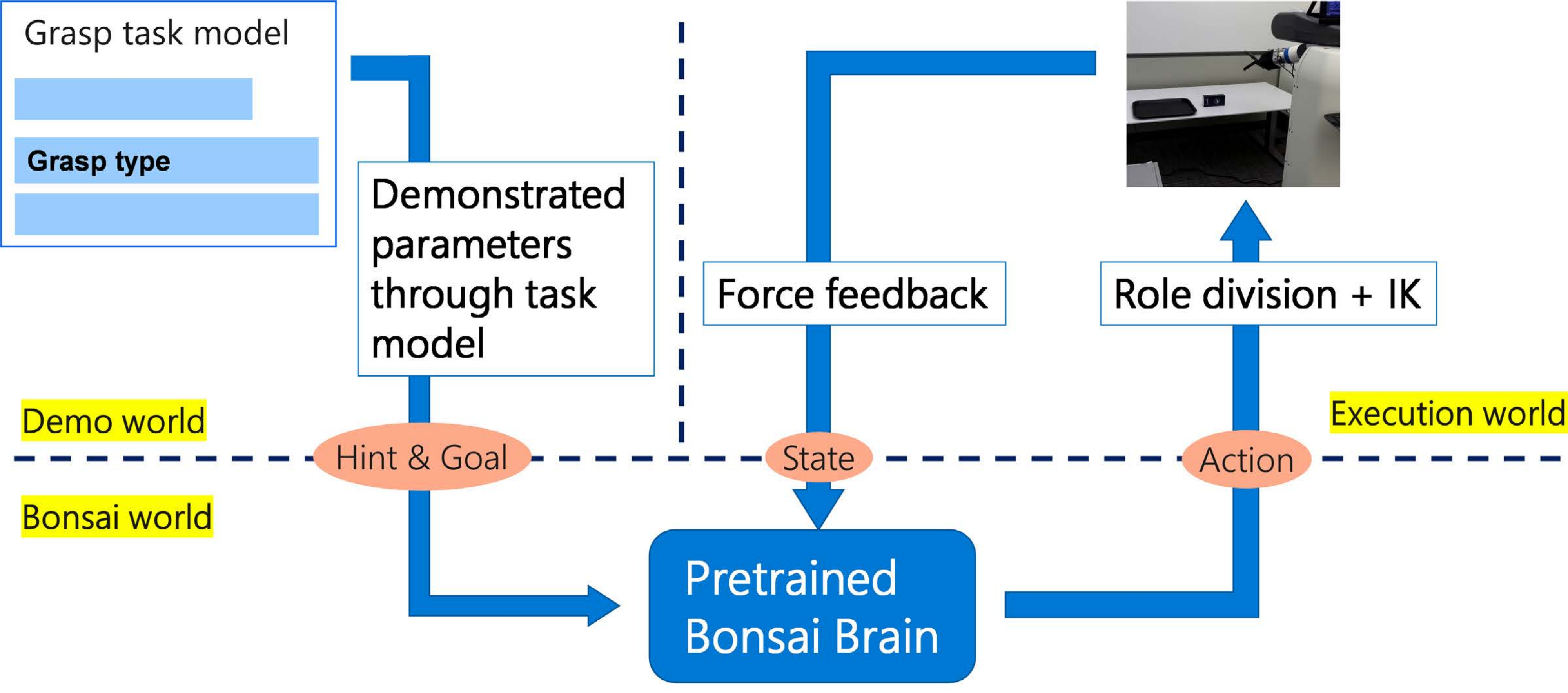}
    \caption{Bonsai brain and parameters}
    \label{fig:bonsai_brain}
\end{figure}

Fig.~\ref{fig:bonsai_brain} shows the summary of the information flow to/from the Bonsai brain performing a grasping task. Each brain is designed to output hand movements using the demonstration parameters as hint values and the forces from the environment as state values. To minimize the differences between Sim2Real and between different sensors, the values from the force sensor are not used directly, but are processed to see if they exceed a certain threshold value. According to the hand position specified by the Bonsai brain, IK is solved by relying on human Labanotation. The robot's overall posture is also determined using the robot's mobile and lifting mechanisms as well to satisfy the required shoulder position required by the role-division algorithm.

\section{Robot execution}

\begin{figure}[h]
\begin{tabular}{cc}

\begin{minipage}{0.5\hsize}
\centering
\includegraphics[height=30mm]{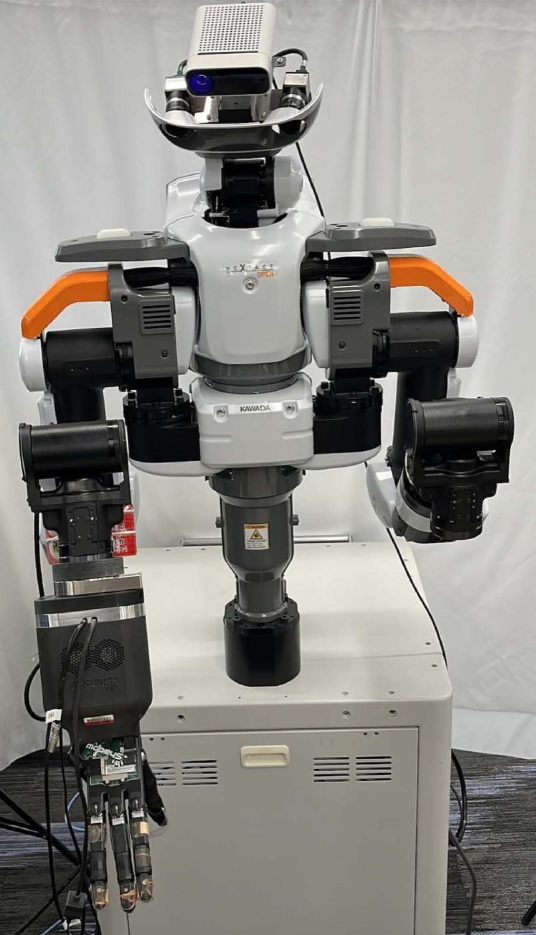} \\
(a) Nextage with Shady
\end{minipage}

&

\begin{minipage}{0.5\hsize}
\centering
\includegraphics[height=30mm]{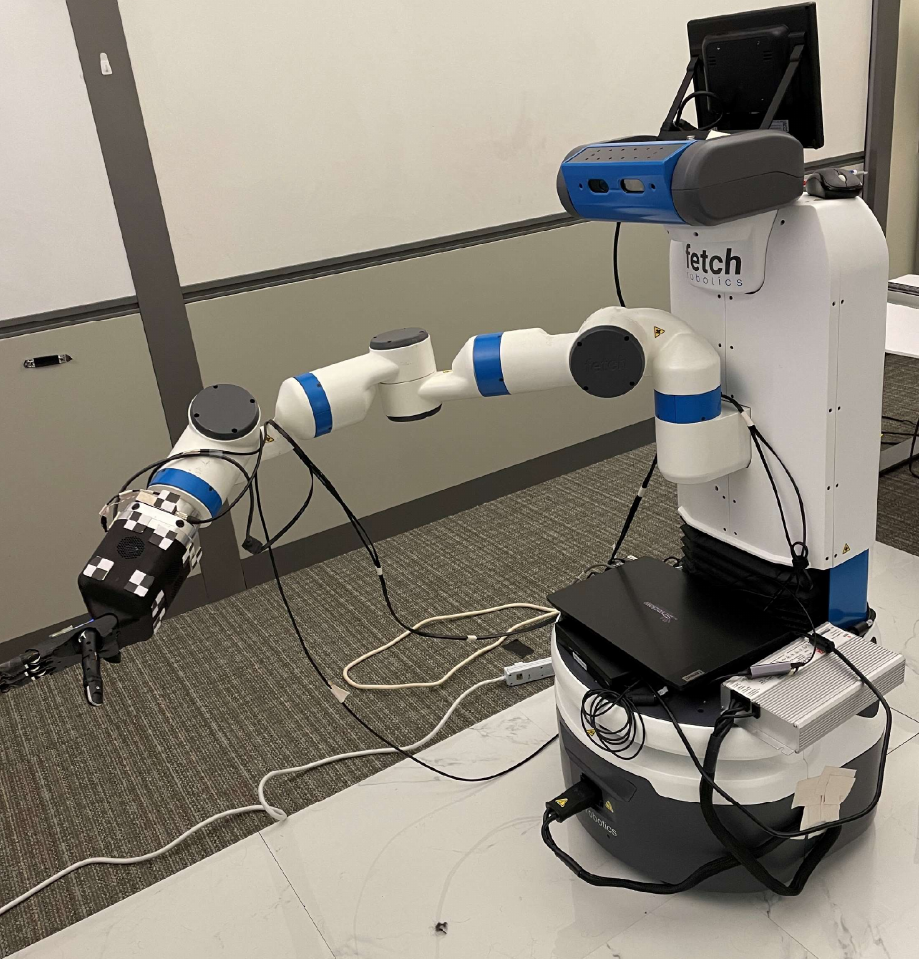} \\ 
(b) Fetch with Shady
\end{minipage}

\end{tabular}

\caption{Test bed.}
\label{fig:testbed}

\end{figure}

To emphasize reusability at the hardware level~\cite{takamatsu2022learning}, two robots, Nextage and Fetch, both running in ROS, were used as testbeds. Nextage by Kawada Robotics has two arms, each with six degrees of freedom, with one degree of freedom (rotation around a vertical axis) at the waist. See Fig.~\ref{fig:testbed}. In this experiment, the robot did not use its left arm or waist, working only with its right arm, which is equipped with a Shadow Dextrous Hand Lite from Shadow Robotics as a robotic hand. Nextage is equipped with a stereo camera to observe the environment. The Fetch Mobile Manipulator has 7 DOFs in the arms with 1 DOF at the waist (vertical movement) and 2 DOFs in the mobile base. The Fetch robot is also equipped with a Shadow Dexterous Hand Lite. It is also equipped with a Primesense Carmine 1.09 RGB-D camera for environmental monitoring. In this paper, as an example of hardware-level reusabiliy, the same task-model sequence, generated from a demonstration, is given to the TSS to control the two robots, which share the same skill libraries, with only different IKs.

\begin{figure}[h]
    \centering
       \vspace{1mm}
    \includegraphics[width=\linewidth]{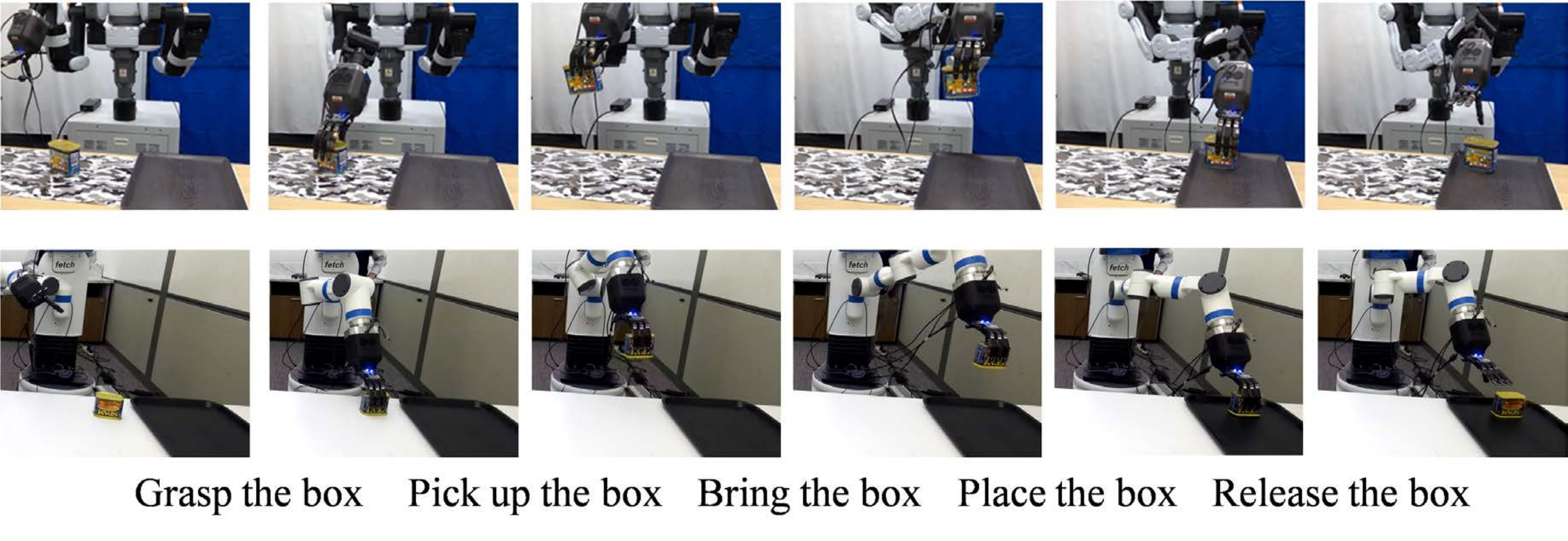}
    \caption{Executions of Box-displacement GMR operation}
    \label{fig:place-on-plate-demo}
\end{figure}

\paragraph{Box-placement GMR operation}
A box-displacement GMR operation is demonstrated in front of Azure Kinect with the verbal instructions, as shown in Fig.~\ref{fig:demonstration}.
The task model sequence, consisting of:
\begin{enumerate}
    \item grasp the box (active-force closure)
    \item pick up the box from the desk (PTG11)
    \item bring-carefully the box (STG12)
    \item place the box on a plate (PTG13), and 
    \item release the box (active-force closure). 
\end{enumerate}
was obtained and initiated from the verbal input, and then, corresponding skill parameters for each task models are obtained from the visual demonstration. 
The task model sequence was uploaded to Azure and, then down loaded to the local TSS at the Shinagawa site controlling Nextage in Shinagawa as shown in the upper raw of Fig.~\ref{fig:place-on-plate-demo}. In the lower raw of Fig.~\ref{fig:place-on-plate-demo}, the same task model sequence on Azure was downloaded to the local TSS at the Redmond site controlling Fetch at Redmond to perform the tasks. These two TSSs share the same skill libraries, and differ only in the IKs to control the robots.
  
\begin{figure}[h]
    \centering
    \includegraphics[width=\linewidth]{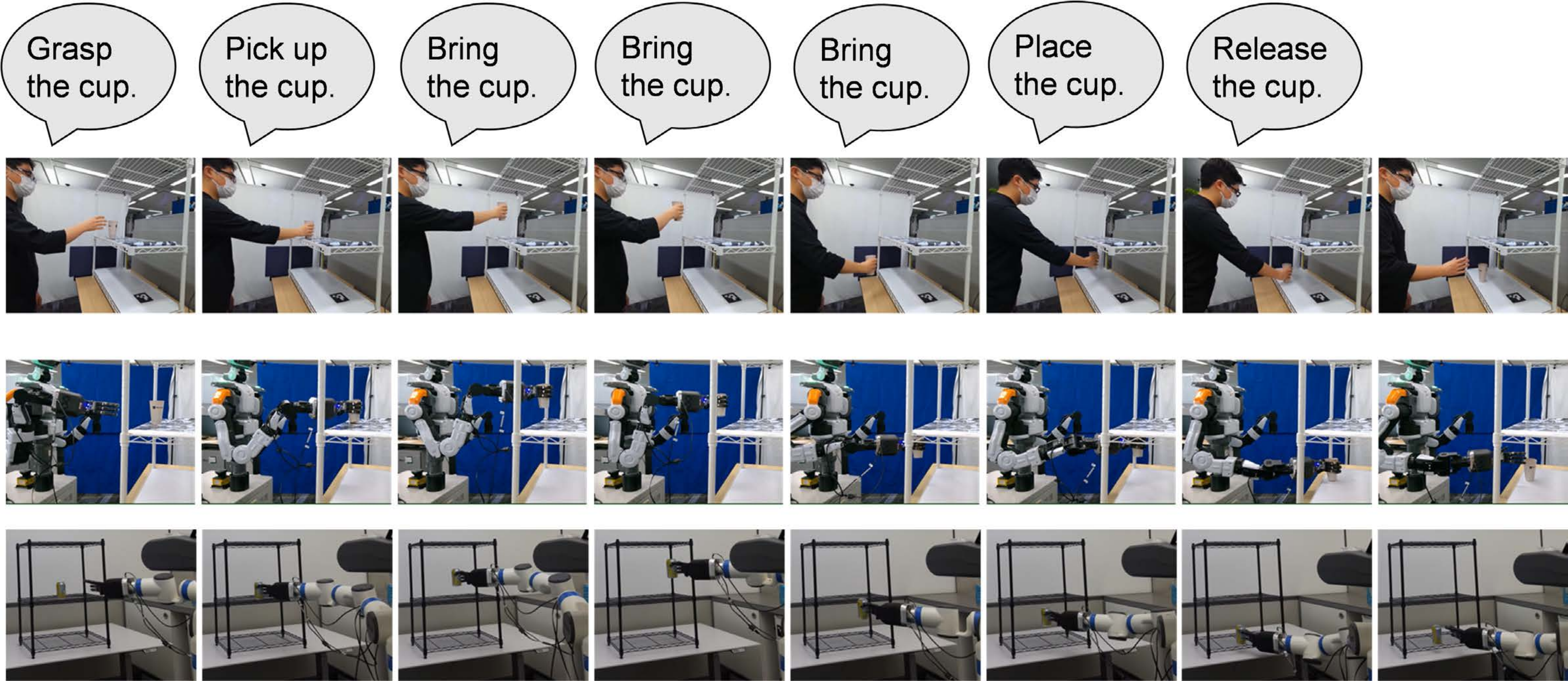}
    \caption{Shelf GMR operation}
    \label{fig:shelf-demo}
\end{figure}

\paragraph{Shelf GMR operation}
The task sequence was identified, from the verbal instruction shown in the upper raw of Fig.~\ref{fig:shelf-demo}, as:
\begin{enumerate}
    \item grasp the cup (Passive-force closure), 
    \item pick up the cup (PTG11),
    \item bring-carefully the cup (STG12), 
    \item bring-carefully the cup (STG12),
    \item bring-carefully the cup (STG12), 
    \item place the cup (PTG13) and 
    \item release the cup (Passive-force closure).
\end{enumerate}
Note that in this operation, bring-carefully STG12 is used three times in a row to teach the robots the trajectory to avoid collision with the shelf. In other words, the demonstrator, not the system, plans the collision avoidance path and teaches the collision avoidance path to the system according to Reddy's 90\% AI rule. See the lower two rows of Fig.~\ref{fig:shelf-demo} for robot's execution.

\begin{figure}[h]
    \centering
    \includegraphics[width=\linewidth]{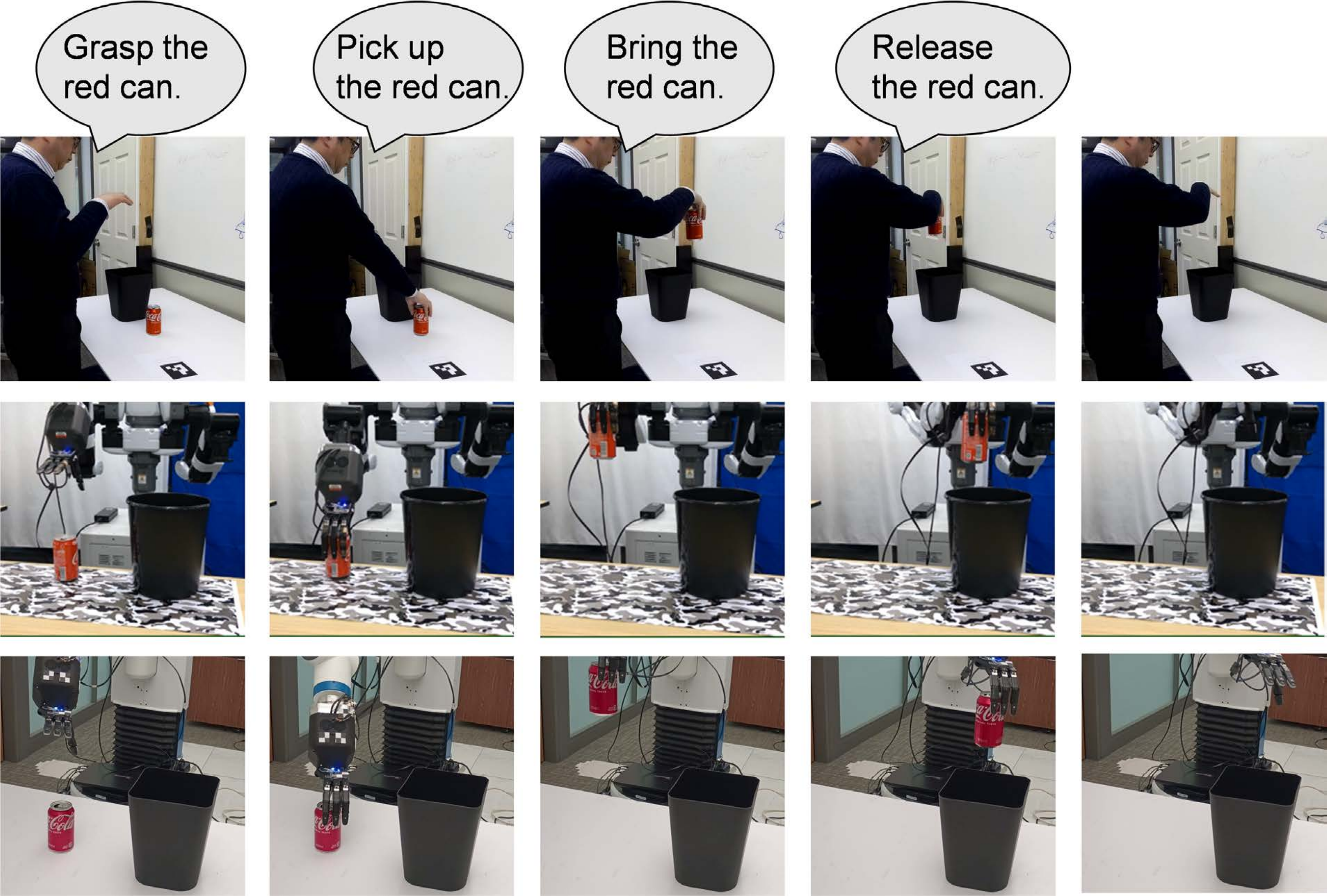}
    \caption{Garbage-disposal GMR operation}
    \label{fig:throw-away}
\end{figure}

\paragraph{Garbage-disposal GMR operation}
The garbage-disposal GMR operation in Fig.~\ref{fig:throw-away} consists of:
\begin{enumerate}
    \item grasp the can (Active-force closure)
    \item pick up the can (PTG11), 
    \item bring the can (STG12), and 
    \item release the can (Active-force closure)
\end{enumerate}. 

The difference between the box-placement operation and this garbage-disposal operation is whether the the object is placed and then released or not placed and then released in the air. It is interesting to note that even if the order of tasks does not change that much, the purpose of the task sequence can change significantly by simply shaving off some of the tasks.

\paragraph{Open-fridge GMR operation}

\begin{figure}[h]
    \centering
    \vspace{1mm}
   \includegraphics[width=0.8\linewidth]{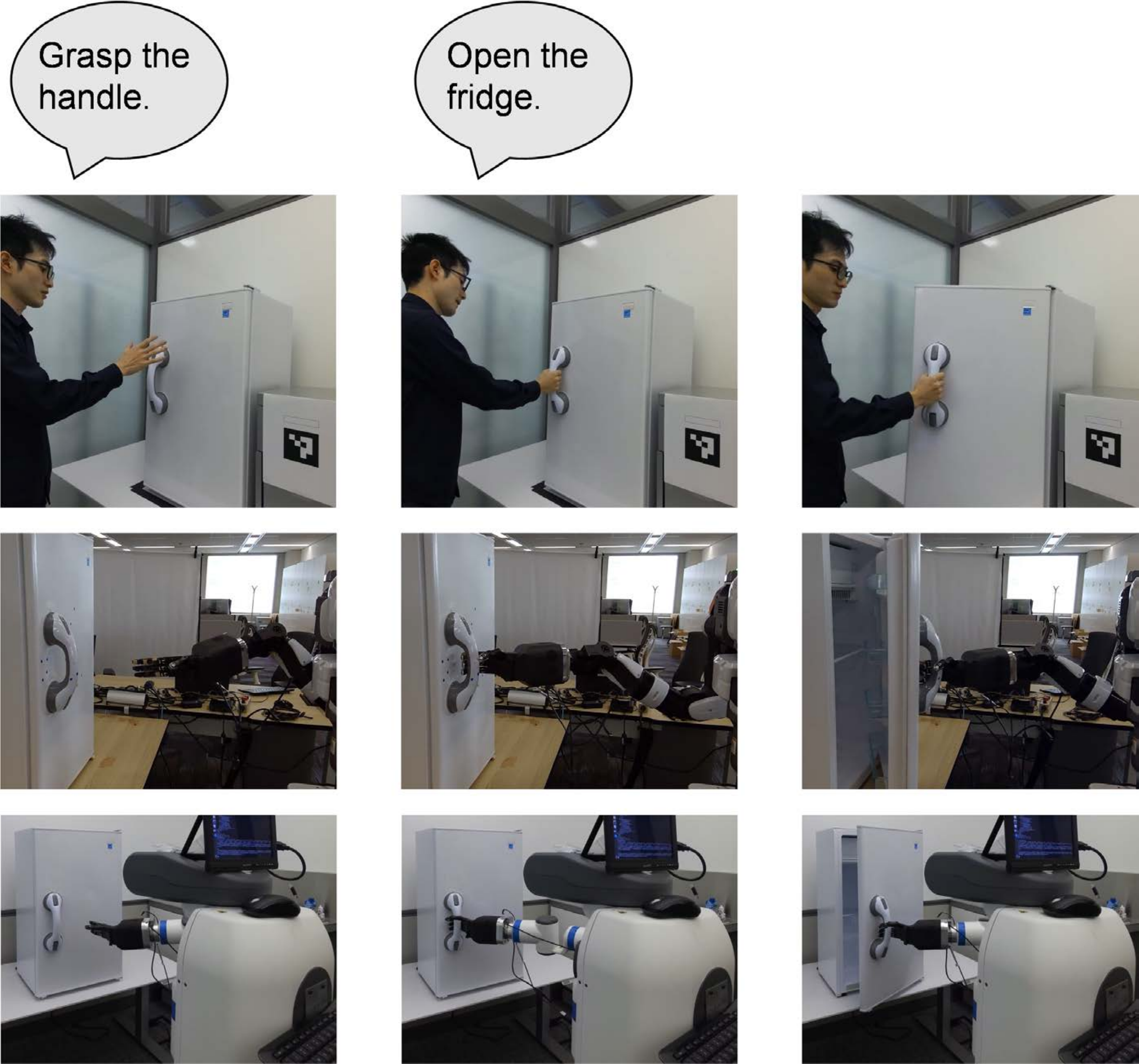}
    \caption{Open-fridge GMR operation}
    \label{fig:open_fridge}
\end{figure}

The Open-fridge GMR operation in Fig.~\ref{fig:open_fridge}  consists of:
\begin{enumerate}
    \item grasp the handle (Loose-closure) and 
    \item open the refrigerator (PTG51).
\end{enumerate}

\section{Summary}

This paper introduces a system for automatically generating robot programs from human demonstrations, referred to as learning-from-observation (LfO), and describes three common sense requirments for applying this system to the household domain. 
Unlike the more common learning-from-demonstration or programming-by-demonstration approaches, LfO does not attempt to directly mimic human trajectories.
Instead, LfO semantically maps human behavior to robot behavior. 
That is, human behavior is first transformed into machine-independent representations, referred to as task models, based on verbal and visual input. 
This representation consists of frameworks that specify what to do, referred to as task models, with associated skill parameters that specify how to do it.
Notably, the skill parameters are not the trajectories themselves, but rather a collection of features obtained from the trajectories and other features needed to perform the task. These task models are then mapped to the actions of each robot. This indirect mapping aims to overcome the kinematic and dynamic differences between humans and robots, and to absorb individual variations in the home environment to which the system is applied.

Unlike conventional industrial domains, the household domain is a cluttered environment and requires the use of human common-sense to guide the system's attention when observing human demonstrations. We focus on common sense with respect to three relationships: the human posture relative to the environment, the grasping strategy of a tool, and the motion of the grasped tool relative to the environment when performing a task sequence. We present the methods for representing them: Labanotation for robot pose, closures and contact webs for grasping, and face-contact transitions and semantic constraints for object motion. The system is designed based on these representations, i.e., task models and skill parameters, as well as recognition and execution systems are designed along these three relations. The system has been applied to actual demonstrations, showing that the generated programs can be executed on robots.

One of the remaining issues is the step-by-step teaching in the task encoder. Currently, a GMR operation is taught verbally with step-by-step instructions at the granularity of tasks, but it would be more convenient to be able to teach a GMR operation in a more general global description. One way is to manually register each GMR operation as a sequence of the tasks that comprise it, and the verbal instruction would be given by the description of the GMR operation. Since the number of GMR operations that commonly occur in the household domain is at most 20 or 30, it would not be too difficult to design the set of necessary GMR operations. Of course, there will still be variations in the verbal input of each GMR operation, which will need to be absorbed by an ML algorithm such as Random Forest. Alternatively, a general description of the GMR operation could be broken down into fine task-level descriptions using an ML algorithm such as chat-GPT. A further possible method would be to hybridize these two methods, using the former registration method for those that often occur frequently and the latter generative method for other exceptional GMR operations that occur, which could be registered as new ones to gradually increase the coverage of registered GMR operations. 

The next challenge is to ground those tasks to those segments in a sequence of a demonstration, assuming that a global description can be decomposed in the granularity of tasks. In particular, in order to determine the skill parameters specific to each task, the segmentation of the video must be accurate and the corresponding segments must be accurately matched to the tasks. Current CNNs for action recognition are too granular in content to be used. A recognition system with fine granuliarity is needed. Alternatively, a generative system such as motion diffusion could be used to generate the actions themselves.

In this system, we considered applying LfO to the household domain. Therefore, it was necessary to introduce the pose representation, the grasping representation and the semantic tasks. Conversely, the number of necessary physical tasks was limited compared to the traditional industrial tasks. In the future, we will consider expanding the number of physical tasks in the library so that the system can be used seamlessly in both of the household and industrial domains.

\bibliographystyle{IEEEtran}
\bibliography{main}

\end{document}